\documentclass[numbers,webpdf,imaiai]{ima-authoring-template}%
\usepackage{amsmath}
\usepackage{amssymb}
\usepackage{mathtools}
\usepackage{thmtools}
\usepackage{bm}
\usepackage{braket}

\graphicspath{{Fig/}}

\theoremstyle{thmstyletwo}%
\newtheorem{theorem}{Theorem}[section]
\newtheorem{proposition}[theorem]{Proposition}%
\newtheorem{lemma}[theorem]{Lemma}%
\newtheorem{corollary}[theorem]{Corollary}%

\newtheorem{example}{Example}%
\newtheorem{remark}{Remark}%

\newtheorem{assumption}{Assumption}

\numberwithin{equation}{section}

\usepackage[capitalize,noabbrev]{cleveref}

\crefname{assumption}{Assumption}{Assumptions}

\renewcommand{\phi}{\varphi}
\renewcommand{\epsilon}{\varepsilon}

\newcommand{\E}{\mathbb{E}} 
\newcommand{\R}{\mathbb{R}} 

\newcommand{\D}{D} 
\newcommand{\cO}{O} 

\newcommand{\cH}[1][t]{\mathcal{H}_{#1}}

\newcommand{\T}{^{\top}} 
\newcommand{\norm}[2][]{\left\Vert #2 \right\Vert_{#1}} 
\newcommand{\abs}[2][]{\left\vert #2 \right\vert_{#1}} 
\newcommand{\inner}[2]{\Braket{#1,#2}} 

\newcommand{\nf}[2]{\nabla F\left(#1;#2\right)}
\newcommand{\nff}[2][S]{\nabla F_{#1}\left(#2\right)}
\newcommand{\nfh}[2]{\nabla^2 F\left(#1;#2\right)}
\newcommand{\nfhf}[2][S]{\nabla^2 F_{#1}\left(#2\right)}
\newcommand{\nphi}[1][\bm{w}]{\nabla_{#1} \Phi}
\newcommand{\hphi}[1][\bm{w}]{\nabla_{#1}^2 \Phi}

\newcommand{\F}[1][S]{F_{#1}}

\newcommand{\x}[1][i]{\bm{x}_{#1}}
\newcommand{\z}[1][i]{\bm{z}_{#1}}
\newcommand{\w}[1][t]{\bm{w}_{#1}}
\newcommand{\wj}[1][i]{\bm{w}^{(#1)}}
\newcommand{\wi}[1][t]{\bm{w}_{#1}^{(i)}}
\newcommand{\tw}[1][t]{\tilde{\bm{w}}_{#1}}
\newcommand{\twi}[1][t]{\tilde{\bm{w}}_{#1}^{(i)}}
\newcommand{\Et}[1][t]{\E_{#1}}

\newcommand{\vv}{\bm{v}}

\newcommand{\Si}[1][i]{S^{(#1)}}

\newcommand{\bi}[2][i]{\beta_{#2}^{#1}}
\newcommand{\Bi}[2][i]{B_{#2}^{#1}}

\newcommand{\poly}{\mathrm{poly}}

\newcommand{\crefassump}{Assumptions~\ref{as:xy},~\ref{as:logistic} and~\ref{as:sigma} }

\begin{document}

\DOI{}
\copyrightyear{}
\vol{}
\pubyear{}
\access{}
\appnotes{Paper}
\copyrightstatement{}
\firstpage{1}


\title[Generalization of SAM
with Interpolating Neural Networks]{Generalization of Sharpness-Aware Minimization with Interpolating Neural Networks}

\author{Tan Sun\ORCID{0009-0005-6640-7684}
\address{\orgdiv{Center of Data Science}, \orgname{ Zhejiang University}, \orgaddress{ \postcode{310000}, \state{Hangzhou}, \country{China}}}}
\author{Junhong Lin*\ORCID{0000-0002-4507-9424}
\address{\orgdiv{Center of Data Science}, \orgname{ Zhejiang University}, \orgaddress{ \postcode{310000}, \state{Hangzhou}, \country{China}}}}

\authormark{Tan Sun and Junhong Lin}

\corresp[*]{Corresponding author: \href{junhong@zju.edu.cn}{junhong@zju.edu.cn}}


\abstract{
	Neural networks often generalize remarkably well even when highly over‑parameterized and perfectly interpolating the training data. 
	Sharpness‑Aware Minimization (SAM) improves generalization by seeking flat minima via a min–max optimization that minimizes loss in a local neighborhood. 
	Despite empirical results, a rigorous theoretical understanding of SAM in interpolating neural networks remains limited. 
	In this work, we study SAM for binary classification with shallow networks under a self‑bounded loss and a neural network realizability assumption, which capture the ability of over‑parameterized networks to interpolate while staying close to initialization.
	Using an algorithmic stability framework, we derive generalization error bounds that avoid the dimension‑dependent factors typical of PAC‑Bayesian analyses. For the full‑batch SAM, we prove that when the network width is sufficiently large, the expected generalization gap is bounded by $O\left(g(1/T)^2/n+1/T\right)$, where $g(\varepsilon)$
	is the parameter deviation from the initialization needed to reach training loss $\varepsilon$, which implies the realizability, $n$ is the sample size, and $T$ the number of iterations. 
	Under the additional NTK‑separability condition and setting 
	$T=\Theta(n)$, this bound simplifies to $\tilde{O}(1/n)$, hiding polylogarithmic factors. We further apply the analysis to non‑interpolating settings, where the bound involves the distance from initialization and the empirical risk of a reference point, and we discuss the analogous results for stochastic SAM and the expected average population loss.
	Our results provide a stability‑based theoretical perspective on SAM’s generalization in over‑parameterized regimes and show that, with sufficient width, SAM achieves favorable sample‑efficient rates while circumventing explicit dependence on the parameter count.	
	}

\maketitle
\allowdisplaybreaks[4]

\section{Introduction}\label{sc:intro}
Neural networks (NNs) have demonstrated outstanding performance in a wide range of machine learning applications. Even when neural networks are highly over-parameterized and perfectly interpolate the training data, they still exhibit strong generalization capabilities. Understanding this remarkable generalization behavior remains a fundamental problem in learning theory. Sharpness-Aware Minimization (SAM) has been introduced as a powerful approach to enhancing generalization \cite{foret2021sharpness}. By formulating a min–max optimization problem that seeks uniformly low loss within a local neighborhood of parameters, SAM effectively identifies flat minima, which are believed to generalize better \cite{keskar2017large,neyshabur2017exploring}. Despite its empirical success, a rigorous theoretical understanding of SAM's generalization properties remains limited, especially for over‑parameterized neural networks in interpolation regime. 

A pioneering generalization bound for SAM has been derived by incorporating the maximal neighborhood loss under Lipschitz continuity and Gaussian perturbation assumptions \cite{foret2021sharpness}.
A Fisher information mask later incorporated into SAM to refine the generalization bound via sparse perturbations, under bounded and smooth gradients assumptions \cite{mi2022make}. 
A signal-to-noise decomposition framework has been proposed for noisy binary classification with over-parameterized ReLU networks; under this framework, SAM is shown to achieve Bayes-optimal error even under weak signal conditions, offering an explanation for its strong performance in noisy settings \cite{chen2023why}. 
The analysis is further refined by assuming a twice‑differentiable loss, explicitly incorporating Hessian terms into the generalization bound and employing curvature regularization \cite{wu2024cr}. 
Existing theoretical analyses remain within the PAC-Bayesian framework \cite{mcallester2003simplified,neyshabur2018PACBayesian} with alternative tools such as Rademacher complexity or algorithmic stability rarely employed. Moreover, these works have not yet fully exploit the intrinsic properties of the neural network loss landscape.

To better understand the generalization of SAM in the interpolation regime, we adopt the NN-realizability assumption, which states the existence of a non-increasing function $g:\R_{+}\rightarrow\R_{+}$ such that for any $\epsilon>0$, there exists a parameter with training loss at most $\epsilon$ and distance from the initialization at most $g(\epsilon)$ \cite{taheri2024generalization}. This property has been explicitly or implicitly validated in numerous theoretical studies of ReLU and tanh networks, covering both the lazy and rich regimes \cite{jacot2018Neural,wu2018How,ji2020polylogarithmic,e2020Machine,e2022Barron,schliserman2022Stability}.
In particular, when the data satisfy the Neural Tangent Kernel (NTK)-separability condition \cite{taheri2024generalization,nitanda2020Gradient,chen2021how,cao2019Generalization}, meaning that, there exists a unit vector that linearly separates the data with margin $\gamma > 0$ under the NTK feature map, it can be shown that $g(\epsilon)=O\left(\mathrm{poly}\log(1/\epsilon)\right)$ \cite{taheri2024generalization}.

In this work, we derive generalization error bounds--both in terms of estimation error and expected population loss--for SAM-trained neural networks in the interpolating regime, under self-bounded loss and realizability assumptions. Unlike prior studies that primarily rely on the PAC-Bayesian framework, we establish generalization guarantees for SAM via algorithmic stability, explicitly leveraging the structural properties of the network. Under the NN-realizability assumption, we prove that for the network width $m\geq\Omega\left(g(\epsilon)^2\right)$, SAM achieves estimation error $O\left(\frac{g(\epsilon)^2}{n}+\frac{1}{T}\right)$ and expected population loss  $O\left(\left(1+\frac{\sqrt{T}}{n}\right)\frac{g(\epsilon)^2}{\eta T}\right)$, where $T, \eta$ and $n$ denote the number of steps, learning rate and training sample size. When the NTK-separability condition further holds (so $g(\epsilon)=O\left(\mathrm{poly}\log(1/\epsilon)\right)$), these bounds reduces to $\tilde{O}\left(\frac{1}{n}\right)$ and $\tilde{O}\left(\frac{1}{\sqrt{n}}\right)$ with appropriate settings of $T$ and $\eta$.
Compared with existing PAC-Bayesian bounds for SAM, our results avoid dependence on the parameter dimension, which often grows rapidly in over‑parameterized settings. 

Our analysis proceeds by establishing bounds under the NN‑realizability assumption for interpolating networks, and then specializing them to the NTK‑separable case. The proof builds on algorithmic stability analysis framework: we first show that the SAM iterates remain bounded and converge and we then relate the generalization error to on-average model stability \cite{lei2020fine}. A key aspect of our analysis lies in the loss landscape properties of the objective function--specifically, the weak convexity and weak Lipschitz continuity induced by the self-boundedness of the objective function as the network width grows.

The remainder of this paper is organized as follows: \cref{sc:pre} presents the problem setup and the primary assumptions. \cref{sc:results} introduces our main theoretical findings regarding the generalization error of SAM. \cref{sc:results2} provides further results for expected population loss. \cref{sc:pf-strategy} presents the proof strategy. \cref{sc:rela} discusses other related works. \cref{sc:con} concludes the paper. All omitted proofs are deferred to the appendices.

\paragraph*{Related works.} 
A brief discussion of related work is deferred to \cref{sc:rela}; it covers variants of SAM, its convergence and generalization analyses, as well as broader generalization theory for neural networks.

\paragraph*{Notations.} 
We denote $[n]=\{1,2,\cdots,n\}$. 
We use bold lowercase letters, e.g., $\w[]$, to denote vectors, capital letters, e.g., $W$ to denote matrices.
We use $\norm{\cdot}$ for the $\ell_2$ norm of vectors and the operator norm of matrices, and $\norm[F]{\cdot}$ for the Frobenius norm of matrices. 
We use $\lambda_{\min}\left(\cdot\right)$ to denote the minimum eigenvalue of a matrix.
We use the standard big-$O$ and big-$\Omega$ notations. We also use $\tilde{O}$ to hide the polylogarithmic factors. 
We denote by $[\w[1],\w[2]]=\{\w[]\in\R^{d}:\w[]=\alpha\w[1]+(1-\alpha)\w[2], \alpha\in[0,1]\}$ the line segment between $\w[1]$ and $\w[2]$ in $\R^{d}$.
For a function $\Phi(\w[],\x[]):\R^{d'}\times\R^{d}\rightarrow\R$, we denote $\nphi(\w[],\x[])$ as the gradient with respect to $\w[]$.

\section{Preliminary}\label{sc:pre}
In this section, we introduce the problem setup. We focus on the binary classification problem with the following empirical and population risks:
\begin{align}
		& \F(\w[]):=\frac{1}{n}\sum_{i=1}^{n}F(\w[];\z),\quad
		\F[\D](\w[]) := \E_{\z[]\sim\D}\left[F(\w[];\z[])\right]. \nonumber
\end{align}
Here, $S = \left\{\z[1],\dots,\z[n]\right\}$ denotes the training set drawn independently from a distribution $\D$, where each sample $\z=(\x,y_i)$ consists of a feature vector $\x\in\R^{d}$ and a label $y_i\in\{\pm1\}$. The loss on a single sample is $F(\w[];\z[])=f\left(y\Phi(\w[],\x[])\right)$. $\Phi(\w[],\x[])$ is the neural network output and $f:\R\rightarrow\R_{+}$ is the loss function, which is assumed to be twice differentiable, non-negative and convex throughout the paper. 
We are interested in the expected generalization error $\E\left[\F[\D]\left(\w[T]\right)-\F\left(\w[T]\right)\right]$ and the expected population loss $\E\left[\F[\D]\left(\w\right)\right]$. 

We now introduce the key assumptions on the data and the model.

\begin{assumption}[Data Distribution]\label{as:xy}
	For $\z[]=(\x[],y)\sim\D$, we have $\norm{\x[]}\le R$ and $y\in\{\pm1\}$ for some positive constant $R>0$.
\end{assumption}
\begin{assumption}[Self-Bounded Loss]\label{as:logistic}
	The loss function $f:\R\rightarrow\R_{+}$ is convex and twice differentiable, satisfying $\abs{f'(u)}\le\beta_1$, $\abs{f''(u)}\le\beta_2$ and $\abs{f'(u)}\le \beta_3f(u)$ ($\beta_3$-self-bounded), for any $u\in\R$ with some positive constants $\beta_1,\beta_2,\beta_3>0$.
\end{assumption}
The logistic loss $f(u)=\log\left(1+e^{-u}\right)$ satisfies \cref{as:logistic} with $\beta_1=1,\beta_2=\frac{1}{4}$ and $\beta_3=1$. 
It is worth noting that although the loss function $f$ is convex, the per‑sample loss $F(\cdot;\z[])$ and the empirical risk $\F(\cdot)$ are generally non-convex due to the network parameterization.

We consider the one-hidden layer neural network $\Phi(\w[],\x[]):\R^{d'}\times\R^{d}\rightarrow\R$ with $m$ hidden neurons: 
\begin{align}\label{eq:s-nn}
	& \Phi(\w[],\x[]) = \frac{1}{\sqrt{m}}\sum_{j=1}^{m}a_j\sigma\left(\inner{\w[]^{(j)}}{\x[]}\right), 
\end{align}
where $\sigma:\R\rightarrow\R$ is the activation function, $\w[]^{(j)}\in\R^{d},j\in[m]$ are the weights of the $j$-th hidden neuron, and $\frac{a_j}{\sqrt{m}}, j\in[m]$ are fixed second-layer weights with $a_j\in\{\pm1\}$. The full parameter vector is  $\w[]=\left[\w[]^{(1)};\dots;\w[]^{(m)}\right]\in\R^{d'}$ with dimension $d'=md$. During the optimization process, we only train the first layer and keep the second layer fixed, following the same setup in e.g., \cite{ji2020polylogarithmic,nitanda2020Gradient,du2019gradient,li2018learning,arora2019fine}.

For the activation function, we make the following assumption.
\begin{assumption}[Activation Function]\label{as:sigma}
	The activation function $\sigma:\R\rightarrow\R$ is twice differentiable, satisfying $\abs{\sigma'(u)}\le G_1$, $\abs{\sigma''(u)}\le G_2$ and $\sigma(0)=\sigma_0$ with constants $G_1, G_2>0$ and $\sigma_0\in\R$.
\end{assumption}
\begin{remark}
	\cref{as:sigma} holds for several typical activation functions such as Sigmoid $\sigma(u)=\frac{1}{1+e^{-u}}$ with $G_1=\frac{1}{4},G_2=\frac{\sqrt{3}}{18}$, Tanh $\sigma(u)=\frac{e^{u}-e^{-u}}{e^{u}+e^{-u}}$ with $G_1=1,G_2=\frac{4\sqrt{3}}{9}$, Softplus $\sigma(u)=\log(1+e^{u})$ with $G_1=1,G_2=\frac{1}{4}$ and GELU $\sigma(u)=\frac{1}{2}u\left(1+\mathrm{erf}\left(\frac{u}{\sqrt{2}}\right)\right)$ with $G_1=\frac{1}{2}\left(\mathrm{erf}(1)+1+\frac{2}{e\sqrt{\pi}}\right),G_2=\frac{2}{\pi}$. 
\end{remark}
For simplicity, we denote $G=\max\left\{1,G_1,G_2\right\}$ throughout the paper.

We now adopt the neural network realizability assumption, which captures the ability of over‑parameterized networks to achieve arbitrarily small training error while remaining close to their initialization.

\begin{assumption}[NN Realizability] \label{as:f-nn-realizability}
	There exists a non-increasing function $g:\R_{+}\rightarrow\R_{+}$ such that for any training set $S=\{\z[1],\dots,\z[n]\}$ drawn independently from $\D$, any sufficiently small $\epsilon>0$ and initialization $\w[1]$, there exists $\w[\epsilon]\in\R^{d'}$ satisfying $\F\left(\w[\epsilon]\right) \le \epsilon$ and $g(\epsilon) \ge \norm{\w[\epsilon]-\w[1]}$.
\end{assumption}
This assumption reflects a well-established expressive power of over-parameterized neural networks: they are capable of interpolating the training data (achieving near-zero empirical loss) with parameters lying in a bounded neighborhood of the initialization. 
Such a property has been demonstrated for ReLU networks, tanh networks, as well as shallow architectures, in various realizable settings \cite{taheri2024generalization,ji2020polylogarithmic,chen2021how,taheri2025sharper}.

In certain parts of the analysis, it is convenient to work with a point‑wise variant of the same realizability assumption, where the loss on every individual sample is required to be small.
\begin{assumption}\label{as:nn-realizability}
	There exists a non-increasing function $g:\R_{+}\rightarrow\R_{+}$ such that for any training set $S=\{\z[1],\dots,\z[n]\}$ drawn independently from $\D$, any sufficiently small $\epsilon>0$ and initialization $\w[1]$, there exists $\w[\epsilon]\in\R^{d'}$ satisfying $F\left(\w[\epsilon];\z\right) \le \epsilon,\ i\in[n]$, and $g(\epsilon) \ge \norm{\w[\epsilon]-\w[1]}$.
\end{assumption}

\Cref{as:nn-realizability} is slightly stronger than \cref{as:f-nn-realizability} because point‑wise small loss directly implies small average loss.
Note that the point-wise loss version directly implies the average loss version. Conversely, by substituting $g(\epsilon/n)$ in place of $g(\epsilon)$, the weaker assumption yields point‑wise guarantees, which does not affect the order up to a $\log n$ factor as the examples shown in \cref{prop:linear,prop:ntk}.

Moreover, the specific form of $g$ depends on the underlying data geometry as illustrated by the following examples. Under the linear separability condition in \cref{prop:linear} with logistic loss and tanh activations, we obtains $g(\epsilon)=\frac{2}{\gamma}\log(1/\epsilon)$. Under the NTK-separability condition in \cref{prop:ntk}, assuming the network width satisfies $m\ge \Omega\left(\log^4(1/\epsilon)\right)$, we may take $g(\epsilon)=\frac{1}{\gamma}\left(2C+\log(1/\epsilon)\right)$, where $C$ bounds the initial network output uniformly. These examples confirm that $g(\epsilon)$ grows only logarithmically in $1/\epsilon$. This slow growth is crucial for our subsequent optimization and generalization analyses: it allows us to control the distance from initialization while still achieving optimization and generalization bounds. We provide the proofs for \cref{prop:linear,prop:ntk} and examples satisfying the NTK-separability in \cref{sc:example}. 

\begin{proposition}[Linear separability]\label{prop:linear}
	Let $\Phi$ be in \cref{eq:s-nn}, $\sigma$ be tanh and $\w[1]=\bm{0}$. Let \cref{as:xy,as:logistic} hold. For any training set $S=\{\z[1],\dots,\z[n]\}$ drawn i.i.d. from $\D$, if there exists a unit vector $\vv^*\in\R^{d}$ such that $y_i\inner{\vv^*}{\x}\ge\gamma>0$ for any $i\in[n]$, then \cref{as:f-nn-realizability,as:nn-realizability} hold with $g(\epsilon)=\frac{2}{\gamma}\log(1/\epsilon)$.
\end{proposition}

\begin{proposition}[NTK-separability]\label{prop:ntk}
	Given any $S$. Let $\Phi$ be in \cref{eq:s-nn} and $\w[1]\in\R^{d'}$ satisfy $\abs{\Phi(\w[1],\x)}\le C$ for any $i\in[n]$. Assume that \crefassump hold. For any training set $S=\{\z[1],\dots,\z[n]\}$ drawn i.i.d. from $\D$, if there exists an unit vector $\w[*]\in\R^{d'}$ such that $y_i\inner{\nphi(\w[1];\x)}{\w[*]}\ge\gamma>0$ for any $i\in[n]$ and $m\ge\Omega\left(\log^4(1/\epsilon)\right)$, then \cref{as:f-nn-realizability,as:nn-realizability} hold with $g(\epsilon)=\frac{1}{\gamma}\left(2C+\log(1/\epsilon)\right)$.
\end{proposition}

\begin{remark}
	The same results continue to hold if the conditions $\F(\w[\epsilon])\le\epsilon$ in \Cref{as:f-nn-realizability} or $F(\w[\epsilon];\z)\le\epsilon$ in \Cref{as:nn-realizability} are weakened to an average over $T$ training examples: $\frac{1}{T}\sum_{t=1}^T F(\w[\epsilon];\z[i_t])\le\epsilon$, where $i_1,\dots,i_T$ are drawn uniformly from $[n]$. Indeed, the concentration inequality ensures that $\frac{1}{T}\sum_t F(\w[\epsilon];\z[i_t]) - \F(\w[\epsilon]) = \mathcal{O}(1/\sqrt{T})$ with high probability, as long as $F(\w[\epsilon];\z[i_t])\le O\left(1\right)$. Hence, taking $T = \Omega(1/\epsilon^2)$ guarantees $\F(\w[\epsilon])\le O(\epsilon)$.
\end{remark}

We focus on the SAM algorithm \cite{foret2021sharpness}, which updates the parameters by considering a perturbation in the direction of the gradient, scaled by a factor $\rho$. The update rule at iteration $t$ is:
\begin{align}
	& \tw = \w + \rho \frac{\nabla F(\w)}{\norm{\nabla F(\w)}}, \nonumber\\
	& \w[t+1] = \w - \eta \nabla F(\tw), \nonumber
\end{align}
where $\nabla F(\cdot)$ represents the gradient. The whole algorithm steps are summarized in \Cref{alg}.

\begin{figure}
	\centering
	\begin{algorithm}[H]
		\caption{SAM}
		\label{alg}
		\begin{algorithmic}
			\Require Iteration number $T$, initialization $\w[1]$, training set $S$, learning rate $\eta>0$, perturbation radius $\rho\geq 0$;
			\For{$t=1,2,\dots,T$}
			\State \textbf{Option 1: Full-batch Gradient}
			\State $\tw = \w + \rho \frac{\nabla\F(\w)}{\norm{\nabla\F(\w)}}$;
			\State $\w[t+1] = \w  - \eta \nabla\F(\tw)$;
			\State \textbf{Option 2: Stochastic Gradient}
			\State Draw a random sample $\z[i_t]$ uniformly from $S$;
			\State $\tw = \w + \rho \frac{\nf{\w}{\z[i_t]}}{\norm{\nf{\w}{\z[i_t]}}}$;
			\State $\w[t+1] = \w  - \eta \nf{\tw}{\z[i_t]}$;
			\EndFor
			\Ensure $\w[1],\dots,\w[T]$.
		\end{algorithmic}
	\end{algorithm}
\end{figure}

\section{Generalization Bounds of SAM for Shallow Neural Network}\label{sc:results}
In this section, we state our main generalization bounds.
Throughout, $\E$ denotes the expectation over the random draw of the training set $S$; in the stochastic setting, it additionally averages over the random indices $\{i_t\}$ used in \Cref{alg}.

\begin{theorem}\label{thm:fs-gen-re}
	Assume $T\ge1$ and $\epsilon\le 1/T$. 
	Let \cref{as:xy,as:logistic,as:sigma,as:f-nn-realizability} hold. 
	Let $\Phi$ be the network in \cref{eq:s-nn} and $S=\{\z[1],\dots,\z[n]\}$ a training set drawn independently from $\D$. 
	Let $\left\{\w\right\}_{t=1}^{T}$ be produced by the full-batch variant of \cref{alg} with initialization $\w[1]$.  
	Define 
	$C_1=GR\max\left\{\beta_1,\sqrt{2\beta_2},\beta_3\right\}, C_2=G^2R\max\left\{\beta_1,\beta_3\right\}$ and $L_F=C_1^2+\frac{C_2}{\sqrt{m}}$. 
	If $m\ge32^2C_2^2g(\epsilon)^4$,  $0<\eta\le\min\left\{\frac{1}{6C_1^2},\frac{1}{4L_F},\frac{\norm{\w[1]-\w[\epsilon]}^2}{8}\right\}$ and $0\le\rho\le\min\left\{\frac{1}{\eta T^2},\frac{\norm{\w[1]-\w[\epsilon]}}{T}\right\}$, then
	\begin{align}
		& \E\left[\F[\D](\w[T])-\F(\w[T])\right] 
		\leq
		O\left(\frac{g(\epsilon)^2}{n} 
		+ \frac{1}{T} \right). \nonumber
	\end{align}
\end{theorem}

This theorem provides a generalization bound for the SAM algorithm in the shallow neural network under the interpolation assumption. 
When the neural network width $m$ is sufficiently large, scaling as $\Omega\left(\mathrm{poly}(g(\epsilon))\right)$, SAM achieves an expected generalization gap of order $O(g(\epsilon)^2/n + 1/T)$. The next theorem removes the interpolation requirement and holds for any reference parameter satisfying a mild condition on its empirical risk.
\begin{theorem}\label{thm:fs-gen}
	Assume $T\ge1$. 
	Let \crefassump hold. 
	Let $\Phi$ be the network in \cref{eq:s-nn} and $S=\{\z[1],\dots,\z[n]\}$ a training set drawn independently from $\D$. 
	Let $\left\{\w\right\}_{t=1}^{T}$ be produced by the full-batch variant of \cref{alg} with initialization $\w[1]$. 
	Let $C_1, C_2$ and $L_F$ be defined as in \cref{thm:fs-gen-re}. 
	Suppose there exists a parameter satisfying 
	\begin{align}\label{eq:f-w}
		\norm{\w[]-\w[1]}^2\ge8\eta T\F(\w[]).
	\end{align} 
	If $m\ge32^2C_2^2\norm{\w[]-\w[1]}^4$,  $0<\eta\le\min\left\{\frac{1}{6C_1^2},\frac{1}{4L_F}\right\}$ and $0\le\rho\le\min\left\{\frac{1}{\eta T^2},\frac{\norm{\w[1]-\w[]}}{T}\right\}$, then we have,
	\begin{align}
		\E\left[\F[\D](\w[T])-\F(\w[T])\right] 
		&\leq
		O\left(\frac{\norm{\w[]-\w[1]}^2}{n}
		+ \frac{1}{T} \right). \nonumber
	\end{align}
\end{theorem}

\Cref{thm:fs-gen} extends the previous bound to non‑interpolating scenarios: the reference parameter $\w[]$ need only satisfy \eqref{eq:f-w}, which balances its distance from initialization with the cumulative empirical loss. The resulting bound naturally involves both $\norm{\w[]-\w[1]}^2$ and $\E\left[\F(\w[])\right]$. The proofs of \Cref{thm:fs-gen-re,thm:fs-gen} are given in \Cref{sc:pf-strategy}.

We further apply the results under the linear separability and the NTK-separability.
\begin{corollary}
	Assume the same conditions of \cref{thm:fs-gen} without \cref{eq:f-w}. If the linear separability or the NTK-separability holds (so that $g(\epsilon)=O(\log(1/\epsilon))$ by \Cref{prop:linear,prop:ntk}), and choosing $T=\Theta(n)$, $\epsilon\le\frac{1}{T}$ and $m\ge\Omega\left(\log^4(n)\right)$ yields $\E\left[\F[\D](\w[T])-\F(\w[T])\right] 
	\leq
	O\left(\frac{\poly(\log(n))}{n} \right)$.
\end{corollary}
The above theorems concern the full‑batch setting. In practice, SAM is often run with stochastic gradients, which introduces additional variance. We next show that essentially the same generalization guarantee carries over to this setting, at the cost of a more stringent step‑size requirement.
\begin{theorem}\label{thm:s-gen}
	Assume $T\ge1$. 
	Let \crefassump hold. 
	Let $\Phi$ be the network in \cref{eq:s-nn} and $S=\{\z[1],\dots,\z[n]\}$ a training set drawn independently from $\D$. 
	Let $\left\{\w\right\}_{t=1}^{T}$ be produced by the stochastic variant of \cref{alg} with initialization $\w[1]$. Let $C_1,C_2$ and $L_F$ be defined as in \cref{thm:fs-gen-re}. Suppose there exists a parameter $\w[]\in\R^{d'}$ satisfying almost surely
	\begin{align}\label{eq:s-w}
		\norm{\w[]-\w[1]}^2 &\ge 8\eta\sum_{t=1}^{T}F(\w[];\z[i_t]),
	\end{align}
	where $\{i_t\}$ are the indices drawn in the algorithm.
	If $m\ge16C_2^2\norm{\w[]-\w[1]}^4$,  $0<\eta\le\min\left\{\frac{1}{6C_1^2},\frac{1}{4L_FT}\right\}$ and $0\le\rho\le\frac{\norm{\w[]-\w[1]}}{T}$,
	then we have,
	\begin{align}
		\E\left[\F[\D](\w[T])-\F(\w[T])\right] 
		&\leq
		O\left(\frac{\norm{\w[]-\w[1]}^2}{n}
		+ \frac{\norm{\w[]-\w[1]}}{T} \right) \nonumber
	\end{align}
	In particular, if $m\ge16C_2^2g(\epsilon)^4$, $\eta\le\frac{\norm{\w[1]-\w[\epsilon]}^2}{8}$ and \Cref{as:nn-realizability} holds with $\epsilon\le 1/T$, then taking $\w[]=\w[\epsilon]$ yields
	\begin{align}
		\E\left[\F[\D](\w[T])-\F(\w[T])\right] 
		\le
		O\left(\frac{g(\epsilon)^2}{n} + \frac{g(\epsilon)}{T}\right). \nonumber
	\end{align}
\end{theorem}
\begin{corollary}
	Under the conditions of \Cref{thm:s-gen}, additionally without \cref{eq:s-w}, and that linear separability or NTK‑separability holds.
	Choosing $T=\Theta(n)$, $\epsilon\le\frac{1}{T}$ and $m\ge\Omega\left(\poly(\log(n))\right)$ yields $\E\left[\F[\D](\w[T])-\F(\w[T])\right] 
	\leq
	O\left(\frac{\poly(\log(n))}{n} \right)$.
\end{corollary}

\Cref{thm:s-gen} shows that the generalization bounds from the full‑batch setting carry over to stochastic SAM, up to a factor of $g(\epsilon)$.
The main difference between the two results lies in the requirement on the step size. In the full-batch setting, a constant step size $\eta = O(1)$ suffices to ensure both optimization stability and generalization. In contrast, \cref{thm:s-gen} requires $\eta = O(1/T)$ in order to control the variance introduced by stochastic gradients. 
Condition \Cref{eq:s-w} is the natural stochastic analogue of \cref{eq:f-w}: instead of the total empirical loss $T\F(\w[])$, it constrains the cumulative per‑sample loss along the actual stochastic trajectory. The proofs of \cref{thm:s-gen} are provided in \cref{sc:pf-s}.

\section{Bounds on the Expected Average Population Loss}\label{sc:results2}
In the previous section, we establish generalization bounds for SAM in the full-batch and stochastic settings, under the definition of the expected generalization error $\E\left[\F[\D](\w[T])-\F(\w[T])\right]$. Nevertheless, we are also interested in the average generalization performance over the entire optimization trajectory. Accordingly, in this section, we study the expected average population loss $\frac{1}{T}\sum_{t=1}^{T}\E\left[\F[\D](\w)\right]$, which provides a more complete characterization of the algorithm's long-term behavior.

\begin{theorem}\label{thm:ff-gen}
	Assume $T\ge1$. 
	Let \crefassump hold. 
	Let $\Phi$ be the network in \cref{eq:s-nn} and $S=\{\z[1],\dots,\z[n]\}$ a training set drawn independently from $\D$. 
	Let $\left\{\w\right\}_{t=1}^{T}$ be produced by the full-batch variant of \cref{alg} with initialization $\w[1]$. 
	Let $C_1, C_2$ and $L_F$ be defined in \cref{thm:fs-gen-re}. 
	Suppose there exists a parameter satisfying \cref{eq:f-w}. 
	If $m\ge32^2C_2^2\norm{\w[]-\w[1]}^4$,  $0<\eta\le\min\left\{\frac{1}{6C_1^2},\frac{1}{4L_F},\frac{1}{C_1L_F\sqrt{T}}\right\}$ and $0\le\rho\le\min\left\{\frac{1}{T},\frac{\norm{\w[1]-\w[]}}{T}\right\}$, 
	then we have,
	\begin{align}
		\frac{1}{T}\sum_{t=1}^{T}\E\left[\F[\D](\w)\right] 
		&\leq
		O\left(\left(1+\frac{\sqrt{T}}{n}\right)\frac{\norm{\w[]-\w[1]}^2}{\eta T} 
		+ \frac{\norm{\w[]-\w[1]}}{\sqrt{T}} \right). \nonumber
	\end{align}
	In particular, if \Cref{as:f-nn-realizability} holds with $\epsilon\le\frac{1}{T}$ and $\eta\le\frac{\norm{\w[1]-\w[\epsilon]}^2}{8}$, then taking $\w[]=\w[\epsilon]$ and assuming $m\ge32^2C_2^2g(\epsilon)^4$ yields
	\begin{align}
		& 
		\frac{1}{T}\sum_{t=1}^{T}\E\left[\F[\D](\w)\right] 
		\leq
		O\left(\left(1+\frac{\sqrt{T}}{n}\right)\frac{g(\epsilon)^2}{\eta T}
		+ \frac{g(\epsilon)}{\sqrt{T}} \right). \nonumber
	\end{align}
\end{theorem}
\begin{theorem}\label{thm:ss-gen}
	Assume $T\ge1$. 
	Let \crefassump hold. 
	Let $\Phi$ be the network in \cref{eq:s-nn} and $S=\{\z[1],\dots,\z[n]\}$ a training set drawn independently from $\D$. 
	Let $\left\{\w\right\}_{t=1}^{T}$ be produced by the stochastic variant of \cref{alg} with initialization $\w[1]$. 
	Let $C_1,C_2$ and $L_F$ be defined in \cref{thm:fs-gen-re}. 
	Suppose there exists a parameter $\w[]\in\R^{d'}$ satisfying \cref{eq:s-w} almost surely.
	If $m\ge\max\left\{16C_2^2\norm{\w[]-\w[1]}^4, 36C_2^2T\right\}$,  $0<\eta\le\min\left\{\frac{1}{6C_1^2},\frac{1}{4\sqrt{T}}, \frac{1}{3\sqrt{2T}L_F}\right\}$ and $0\le\rho\le\frac{\norm{\w[]-\w[1]}}{T}$,
	then we have,
	\begin{align}
		\frac{1}{T}\sum_{t=1}^{T}\E\left[\F[\D](\w)\right] 
		&\leq
		O\left(\left(\frac{1}{T}+\frac{1}{Tn}+\frac{1}{n^2}\right)\frac{\norm{\w[]-\w[1]}^2}{\eta}\right) \nonumber
	\end{align}
	In particular, if \cref{as:nn-realizability} holds with $\epsilon\le\frac{1}{T}$ and  $\eta\le\frac{\norm{\w[1]-\w[\epsilon]}^2}{8}$, then taking $\w[]=\w[\epsilon]$ and assuming $m\ge\max\left\{4C_2^2g(\epsilon)^4,6C_2T\right\}$ yields 
	\begin{align}
		\frac{1}{T}\sum_{t=1}^{T}\E\left[\F[\D](\w)\right] 
		\le
		O\left(\left(\frac{1}{T}+\frac{1}{Tn}+\frac{1}{n^2}\right)\frac{g(\epsilon)^2}{\eta}\right). \nonumber
	\end{align}
\end{theorem}

\Cref{thm:ff-gen,thm:ss-gen} characterize the expected average population loss. 
Both theorems require a diminishing step size $\eta = O(1/\sqrt{T})$, in contrast to the constant step size allowed for the generalization gap in \Cref{sc:results}. This is because controlling the average loss along the entire trajectory demands stricter stability conditions. 
In the full‑batch case, the width requirement remains polynomial in $\norm{\w[1]-\w[]}$ as before; the stochastic case additionally requires $m\ge\Omega(T)$ to suppress the gradient variance over $T$ steps.
When $T = \Theta(n)$, the factor $1 + T/n + T^2/n^2$ is constant, and the bound simplifies to $\tilde{O}(1/\sqrt{T})$, i.e., $\tilde{O}(1/\sqrt{n})$.
The proofs of \Cref{thm:ff-gen,thm:ss-gen} are deferred to \Cref{sc:pf-f,sc:pf-s}, respectively.

\section{Proof Strategy and Discussion}\label{sc:pf-strategy}
In this section, we summarize the proof strategy of \cref{thm:fs-gen-re,thm:fs-gen} and discuss the the main implications of our results. The proof combines a
characterization of the local geometric properties of neural networks with a stability-based generalization analysis.

\subsection{Properties of Shallow Neural Networks}
We begin by establishing key properties of the network and the resulting objective function. The formal proofs of all lemmas in this subsection are deferred to \Cref{sc:pf-properties}.
We first characterize the gradient and Hessian of the neural network output with respect to the parameters, which plays a central role in both the optimization and the stability analysis.

\begin{lemma}\label{lem:s-g-h}
	Let $R_1 = GR$ and $R_2=G^2R$. Let \cref{as:xy,as:sigma} hold, then for any $\w[]\in\R^{d'}$,
	\begin{align}
		\norm{\nphi(\w[],\x[])} \le R_1, \quad
		\norm{\hphi(\w[],\x[])} \le \frac{R_2}{\sqrt{m}}. \nonumber
	\end{align}
\end{lemma}	

Based on these bounds, we derive the corresponding properties of the empirical risk $\F(\cdot)$ and per-sample loss $F(\cdot;\z[])$. 
To simplify notation, in this subsection we write $F(\w[])$ to stand for either $\F(\w[])$ or $F(\w[];\z[])$ for any dataset $S$ and sample $\z[]\in\R^d\times\{\pm1\}$ in this subsection.

\begin{lemma}\label{lem:s-weakly-properties}
	Assume the same conditions as in \cref{lem:s-g-h}. If \cref{as:logistic} holds, then for any $\w[]\in\R^{d'}$,
	\begin{align}
		& \norm{\nabla F(\w[])} \le C_1\min\left\{1, \sqrt{F(\w[])}, F(\w[])\right\}, \ 
		\norm{\nabla^2 F(\w[])} \le L_F, \  \lambda_{\min}\left(\nabla^2 F(\w[])\right) \ge -\frac{C_2}{\sqrt{m}}\min\left\{1, F(\w[])\right\}, \nonumber
	\end{align}
	where $C_1=R_1\max\left\{\beta_1,\sqrt{2\beta_2},\beta_3\right\}$, $C_2=R_2\max\left\{\beta_1,\beta_3\right\}$ and $L_F=C_1^2+\frac{C_2}{\sqrt{m}}$.
\end{lemma}

\cref{lem:s-weakly-properties} indicates the self-bounded gradient property and the weakly-convex-like property of $F$, which is central technical lemma in our analysis. Consequently, the objective function behaves increasingly like a convex and Lipschitz-continuous function in the wide network regime. Similar observations have been derived in \cite{taheri2024generalization,taheri2025sharper,sun2026Convergence}.
The properties in \Cref{lem:s-weakly-properties} directly imply the following two lemmas.

\begin{lemma}\label{lem:s-weakly-convex}
	Assume the same conditions as in \cref{lem:s-g-h}. We have
	\begin{align}
		F(\w[]') 
		&\ge F(\w[]) 
		+ \inner{\nabla F(\w[])}{\w[]'-\w[]}
		- \frac{C_2}{2\sqrt{m}}\norm{\w[]'-\w[]}^2. \nonumber
	\end{align}
	Further, if $\norm{\w[]-\w[]'}\le D$ for $D>0$, then
	\begin{align}
		F(\w[]')
		&\ge  F(\w[]) 
		+ \inner{\nabla F(\w[])}{\w[]'-\w[]}
		- \frac{C_2D^2}{2\sqrt{m}} \max_{\vv\in[\w[],\w[]']}  F(\vv). \nonumber
	\end{align} 
\end{lemma}
\begin{lemma}\label{lem:s-max}
	Let $D>0$. Assume the same conditions as in \cref{lem:s-g-h}. Let $\w[],\w[]'\in\R^{d'}$ satisfy that $\norm{\w[]-\w[]'}\le D$. If $\sqrt{m}\ge2C_2D^2$, then for any dataset $S$,
	\begin{align}
		\max_{\vv\in[\w[],\w[]']}  F(\vv) &\le
		\frac{4}{3} \max\left\{ F(\w[]), F(\w[]')\right\}.\nonumber
	\end{align}
\end{lemma}

Together, these lemmas characterize the local geometry of the objective function in the parameter space, which is critical for both the convergence and the generalization analysis.

\subsection{Proof of \cref{thm:fs-gen-re,thm:fs-gen}}
We now outline the proofs of \Cref{thm:fs-gen-re,thm:fs-gen}, which rely on an algorithmic stability argument. The complete proofs are given in \Cref{sc:pf-f}.
The proof proceeds in three steps: (i) bounding the optimization trajectory via induction, (ii) translating this into an on-average stability estimate, and (iii) using stability to bound the generalization gap.

The first step is to establish that the SAM iterates stay within a bounded region around the reference point. This is achieved by an induction argument, leading to the following convergence estimate.

\begin{proposition}[Convergence Bounds]\label{thm:fs-conv}
	Under the same conditions as in \cref{thm:fs-gen}, we have for any $t\in[T]$,
	\begin{align}
		&\sum_{s=1}^{t} \F(\w) \le \frac{4\norm{\w[1]-\w[]}^2}{\eta}. \nonumber
	\end{align}
	Additionally, we have $\norm{\w-\w[]}\le2\norm{\w[1]-\w[]}$ holds for any $t\in[T]$.
\end{proposition}

\cref{thm:fs-conv} shows that the cumulative loss along the trajectory is controlled by the distance from a reference point and the loss at that point. 
Since $F(\w[];\z[])\ge0$ for any $\w[]$ and $\z[]$, we immediately have
\begin{align}
	\norm{\w[]-\w[1]}^2 \ge 8\eta T\F(\w[]) \ge 8\eta T\F[\Si](\w[]), \nonumber
\end{align}
which indicates that $\wi$ and $\F[\Si]$ also satisfy the condition of \cref{thm:fs-conv}.
Thus, 
\begin{align}
	\sum_{t=1}^{T}\F[\Si](\wi) \le \frac{4\norm{\wi[T]-\w[]}^2}{\eta} + \frac{8T}{3}\F[\Si](\w[]), \quad
	\norm{\wi[T]-\w[]} \le 2\norm{\w[1]-\w[]},\ \forall i\in[n]. \nonumber
\end{align}
As a result, we obtain that 
\begin{align}
	\sqrt{m} &\ge 32C_2\norm{\w[1]-\w[]}^2 \ge 4C_2\max_{i\in[n]}\left\{\norm{\w[T]-\w[]}^2+ \norm{\wi[T]-\w[]}^2\right\} \nonumber\\
	&\ge \eta C_2\max_{i\in[n]}\left\{\sum_{t=1}^{T}\F(\w)+ \sum_{t=1}^{T}\F[\Si](\wi)\right\}. \nonumber
\end{align}
With the convergence estimate, we can now bound the on‑average model stability:
\begin{proposition}[Stability Bounds]\label{thm:fs-stability}
	Assume the same conditions as in \cref{thm:fs-gen}. For any $i\in[n]$, let $\Si=\left\{\z[1],\dots,\z[i-1],\z[i+1],\dots,\z[n]\right\}$ be the set
	formed from $S$ by removing the $i$-th element and $\left\{\wi\right\}_{t=1}^{T}$ be produced by the full-batch variant of \cref{alg} with $\F(\w[])$ replaced by $\F[\Si](\w[])$, where $\F[\Si](\w[])=\frac{1}{n}\sum_{j\ne i}F(\w[];\z[j])$.
	If $\sqrt{m}\ge\frac{4\eta C_2}{3}\max_{i\in[n]}\left\{\sum_{t=1}^{T}\F(\w)+ \sum_{t=1}^{T}\F[\Si]\left(\wi\right)\right\}$, then we have for any $t\in[T]$,
	\begin{align}
		\frac{1}{n}\sum_{i=1}^{n}\E\left[\norm{\w[t+1]-\wi[t+1]}\right] &\le \frac{e\eta C_1}{n}\sum_{s=1}^{t}\E\left[\F(\w)\right] + 2e\eta\rho L_Ft. \nonumber
	\end{align}
\end{proposition}
\cref{thm:fs-stability} shows that the expected parameter deviation between the two trajectories grows at most linearly in the accumulated loss. The proofs of \Cref{thm:fs-conv,thm:fs-stability} are deferred to \cref{sc:pf-f}. We now connect the generalization error to on-average model stability:
\begin{align}
	\E\left[\F[\D](\w)-\F(\w)\right]
	&= 
	\frac{1}{n}\sum_{i=1}^{n}\E\left[F\left(\w;\z[]\right)-F\left(\w;\z\right)\right] \nonumber\\
	&= 
	\frac{1}{n}\sum_{i=1}^{n}\E\left[F(\w;\z[])-F\left(\wi;\z[]\right)\right]
	+ 
	\frac{1}{n}\sum_{i=1}^{n}\E\left[F\left(\wi;\z\right)-F\left(\w;\z\right)\right] \nonumber\\
	&\le 
	\frac{2C_1}{n}\sum_{i=1}^{n}\E\left[\norm{\wi-\w}\right], \nonumber
\end{align}
where the last step uses the gradient bound from \Cref{lem:s-weakly-properties}.
Combining this with \Cref{thm:fs-stability} gives
\begin{align}\label{eq:fs-1}
	\E\left[\F[\D]\left(\w\right)-\F(\w)\right]
	&\le \frac{2e\eta C_1^2}{n}\sum_{s=1}^{t-1}\E\left[\F(\w[s])\right] + 4e\eta\rho C_1L_Ft \nonumber\\
	&\le 
	\frac{8e C_1\norm{\w[1]-\w[]}^2}{n} 
	+ 4e\eta\rho L_FT, 
\end{align}
where the final inequality follows from \Cref{thm:fs-conv}. Using the condition, we obtain exactly \Cref{thm:fs-gen}.

Finally, if \cref{as:f-nn-realizability} holds with sufficiently small $\epsilon\leq1/T$, we may take $\w[]$ as $\w[\epsilon]$, for which $\F(\w[\epsilon])\le\epsilon$ and $\norm{\w[1]-\w[\epsilon]}=g(\epsilon)$. The step size condition $\eta \le \frac{\norm{\w[1]-\w[\epsilon]}^2}{8}$ then implies
\begin{align}
	8\eta T\F(\w[\epsilon]) \le T \epsilon \norm{\w[1]-\w[\epsilon]}^2  \le\norm{\w[1]-\w[\epsilon]}^2, \nonumber
\end{align}
so \cref{eq:f-w} is satisfied. Substituting these estimates into the previous bound yields
\begin{align}
	\E\left[\F[\D]\left(\w[T]\right)-\F(\w[T])\right]
	&\le 
	O\left(\frac{g(\epsilon)^2}{n}
	+ \frac{1}{T}\right), \nonumber
\end{align}
which completes the proof \Cref{thm:fs-gen-re}.

The proof of the stochastic variant (\Cref{thm:s-gen}) follows the same stability-based structure as the full-batch case, but requires additional care to handle the sampling variance, relying on conditional expectations rather than the direct expansiveness lemma, and leads to a stricter step size $\eta = O(1/T)$.The bounds on the average population loss (\Cref{sc:results2}) differ principally in that the metric is averaged over the optimization path rather than evaluated at the final iterate, which introduces an extra variance accumulation factor and demands $\eta = O(1/\sqrt{T})$. 

\section{Related works}\label{sc:rela}
We briefly review the most closely related works due to space and knowledge constraints.
\subsection{Sharpness-Aware Minimization}
\paragraph*{Algorithmic Variants of SAM.}
The original SAM algorithm uses a single gradient-ascent step to find flat minima for improved generalization \cite{foret2021sharpness}. Numerous variants of SAM have been proposed to enhance its efficiency, stability, or generalization performance. 
Adaptive SAM (ASAM) adapts the perturbation radius for scale-invariant learning \cite{kwon2021asam}, while Surrogate Gap Guided SAM (GSAM) reduces computational cost overhead via efficient sharpness estimation \cite{zhuang2022surrogate}. LookSAM and Lookahead-SAM leverage historical gradients for periodical SAM steps \cite{liu2022towards,yu2024improving}. Sparse SAM (SSAM) applies binary masks for perturbation sparsity \cite{mi2022make}, and Sharpness-Aware Training for Free (SAF) uses free perturbation approximations \cite{du2022sharpness}. 
Several works tailor SAM to specific regimes or tasks. Gradient-Strength based SAM (GA-SAM) adapts SAM to Natural Language Processing (NLP) tasks \cite{zhang2022ga}; AdaSAM incorporates adaptive learning rate and momentum for faster convergence \cite{sun2024adasam}; and Friendly SAM (FSAM) refines perturbations through gradient decomposition \cite{li2024friendly}. SAM with Adaptive Regularization (SAMAR) applies adaptive regularization, effective for signal processing tasks \cite{zou2025sharpness}.
To enhance stability, Random SAM (RSAM) combines SAM  with random smoothing \cite{liu2022random} and Variance-Suppressed Sharpness-Aware Optimization (VaSSO) suppresses gradient variance \cite{li2023enhancing}. Curvature Regularized SAM (CR-SAM) and Gradient Centralized SAM (GC-SAM) respectively incorporates curvature and gradient centralization for robustness \cite{wu2024cr,hassan2025gcsam}.
For large-scale training, SAMPa parallelizes the gradient computation \cite{xie2024sampa}, Decentralized Stochastic SAM (D-SSAM) implements a distributed variant \cite{chen2024decentralized} and Balancedness-Aware Regularization (BAR) applies to LoRA-style fine-tuning with substantial computational savings \cite{li2024implicit}.
Riemannian SAM extends the idea to manifold spaces \cite{yun2023riemannian}.
A uniform framework including Frob-SAM and Det-SAM handles various invariances and tasks within a single approach \cite{tahmasebi2024universal}.
Finally, $\nu$-SAM employs nuclear norm constraints for memory efficiency \cite{pethick2025sam}.

\paragraph*{Convergence and Generalization Analysis of SAM.}
Theoretical studies of SAM mainly follow two lines: convergence analysis and generalization analysis. On the convergence side, early work reveals that SAM with a constant learning rate may fail to converge under a worst-case perturbation model \cite{andriushchenko2022towards}.
Convergence of the unnormalized SAM to local minima with local acceleration properties has been established, which revels the relationship between convergence and saddle-point escape \cite{sun2018saddle}.
Subsequent research establishes fundamental convergence guarantees for SAM in non-convex optimization, proving sub-linear convergence rates \cite{khanh2024fundamental}, while another study provides a uniform framework and derives improved convergence rates over prior bounds \cite{oikonomou2025sharpness}. Further analysis of SAM dynamics reveals oscillations along high-curvature directions, offering a mechanistic explanation for its bias toward flat minima \cite{bartlett2023dynamics}.
Analysis of SAM with a constant perturbation radius $\rho$ shows that it converges for deterministic smooth convex problems but incurs an unavoidable $\rho^2$ error floor in non-convex deterministic settings and all stochastic variants \cite{si2023practical}.
More recently, leveraging neural-network-specific self‑boundedness and interpolation assumptions, SAM has been shown to achieve an $\tilde{O}(1/T)$ training loss convergence rate for sufficiently wide shallow and deep networks, without requiring bounded gradients or convexity \cite{sun2026Convergence}.
On the generalization side, theoretical bounds demonstrate that in over-parameterized ReLU networks, SAM reduces sharpness more effectively than SGD, leading to improved generalization \cite{chen2023why}. Other work shows that over-parameterization amplifies SAM’s ability to find flat minima, supported by both theoretical and empirical evidence \cite{shin2023effects}.
Recent work has also examined the limitations of SAM’s generalization: while SAM can improve zero-shot out-of-distribution performance, its effectiveness strongly depends on the specific variant, and regret-based analyses in interpolating regimes further compare SAM-like methods with adaptive optimizers from an optimization-theoretic viewpoint \cite{schapiro2024towards}.

\subsection{Generalization in Neural Networks}
\paragraph*{Interpolation and Generalization.}
Modern neural networks exhibit extraordinary generalization capability under over-parameterization and interpolating training. Research indicates that GD exhibits an implicit bias toward solutions with large margin, low rank, or small norm, which enhances generalization \cite{soudry2018separable,lyu2020margin,gunasekar2017matrix,arora2019matrix}. Meanwhile, the network architecture exhibits a natural bias towards simple functions, which facilitates good generalization for the majority of interpolation solutions \cite{perez2019simple,theisen2021abundant,buzaglo2024uniform}. 
Sample complexity bounds that are independent of network size have been derived through the control of metrics such as spectral and Frobenius norm, which further strengthens the theoretical support for implicit regularization behavior \cite{bartlett2017spectral,golowich2018sizeindependent}.

Recent theoretical work on generalization under interpolation has followed several complementary directions. The PAC-Bayesian approach relates generalization bounds to the NTK spectrum or Hessian curvature in the zero-training-error limit \cite{hodgkinson2023PACBayesian,huang2023Analyzing,dziugaite2017Computing}. Local Rademacher complexity analyses for two-layer over-parameterized models yield rates that can surpass standard NTK predictions \cite{yang2025sharp}. Algorithmic stability analyses  show that in realizable or interpolating settings, the stability of GD or SGD leads to fast decay of the risk \cite{taheri2024generalization,taheri2025sharper,lei2022stability}. Results in the NTK or lazy-training regime indicate that in the infinite-width or kernel limit, generalization under interpolation is governed by the kernel spectrum and label noise, thereby permitting benign over fitting \cite{jacot2018Neural,zhu2023benign,arora2019Exact}. 

\paragraph*{Algorithmic Stability in Generalization Analysis.}
Algorithmic stability shows that stable algorithms admit exponential generalization bounds \cite{bousquet2002Stability}.  The framework is later extended to randomized learning algorithms, where non-asymptotic bounds linking random stability to expected generalization error are derived \cite{elisseeff2005stability}.
In the context of optimization, the first stability-based analysis of SGD generalization appears in \cite{hardt2016train}, while subsequent work proposes data-dependent and on-average stability notions, relaxing smoothness and bounded-gradient assumptions and achieving faster generalization rates under low-noise conditions \cite{kuzborskij2018data,lei2020fine}. 
It is also shown that SGD remains stable for non-smooth convex losses, yielding the first dimension-independent generalization bounds for multi-pass SGD \cite{bassily2020stability}. The stability framework is subsequently extended more complex settings, including pairwise learning and minimax problems \cite{lei2021generalization,yang2021simple}. The first stability analysis of adversarial training is also provided in \cite{xing2021algorithmic}. More recent work has substantially sharpened high-probability and moment-based generalization bounds for uniformly stable algorithms, achieving near-optimality via matching lower bounds \cite{feldman2018generalization,bousquet2020sharper}.
Recently, a unified framework has been developed for studying the stability and generalization of biased stochastic gradient methods, establishing on-average model stability bounds for Zeroth-order SGD and Clipped-SGD \cite{zeng2026Stochastic}.

\section{Conclusion}\label{sc:con}
In this paper, we analyze the generalization performance of SAM for neural networks in the interpolation regime. By leveraging algorithmic stability and neural network realizability, we establish generalization bounds that hold with sufficiently wide neural networks. We further extend the analysis to non-interpolating scenarios and specialize the bounds under the NTK-separability condition. 
Our work has several limitations. First, it relies on assumptions such as the self-bounded loss and neural network realizability, which may not hold in all practical scenarios. Second, the theory focuses on fully-connected networks and does not yet cover modern architectures with attention or residual connections. Third, our bounds do not yet fully explain why SAM often outperforms GD or SGD in practice. Moreover, because the study is primarily theoretical, we have not conducted numerical experiments. Addressing these limitations--extending the analysis to richer architectures, incorporating other loss landscapes, and providing empirical validation--represents important directions for future work.
Overall, this work offers a rigorous stability‑based perspective on SAM’s generalization in the interpolation regime.


\appendix

\section{Examples of the Realizability Assumptions}\label{sc:example}

We present examples of data distribution from \cite{taheri2024generalization} satisfying the NTK-separability condition, which is related to the noisy XOR data distribution \cite{wei2019regularization}, a uniform distribution over the following points:
\begin{align}\label{eq:wei}
	& \x[]=\left(x_j\right)_{j=1}^{d}\in\R^{d},\ y\in\R, \quad \mbox{where} \quad
	(x_1,x_2,y)\times(x_3,\dots,x_d) \in \nonumber\\
	& \left\{\left(\frac{1}{\sqrt{d-1}},0,1\right),\left(0,\frac{1}{\sqrt{d-1}},-1\right),\left(\frac{-1}{\sqrt{d-1}},0,1\right),\left(0,\frac{-1}{\sqrt{d-1}},-1\right)\right\}\times
	\left\{\frac{-1}{\sqrt{d-1}},\frac{1}{\sqrt{d-1}}\right\}^{d-2},
\end{align}
and $\times$ denotes the Cartesian product. The following results show that NTK-separability is satisfied when $\{(\x,y_i)\}_{i\in[n]}$ is sampled from this data distribution.

\begin{example}[Proposition 5.3, \cite{ji2020polylogarithmic}]
	Define the Hilbert space $\cH:=\left\{v:\R^{d'}\rightarrow\R^{d'}\mid\int\norm{v(\z[])}^2d\mu_{\mathcal{N}}<\infty\right\}$, where $\mu_{\mathcal{N}}$ is the Gaussian measure. If $d>3$, then for any $(\x[],y)$ drawn from \cref{eq:wei}, there exists $\w[*]\in\cH$ such that $y\inner{\nphi(\w[1],\x[])}{\w[*]}_{\cH}\ge\frac{1}{60d}$.
\end{example}

\begin{example}[Proposition 4.2, \cite{taheri2024generalization}]
	If $\sigma$ is $G$-Lipschitz continuous, $\mu$-strongly convex over the interval $[-2,2]$, each entry of $\w[1]$ is i.i.d. from $\mathcal{N}(0,1)$ and $m\ge\frac{80^2d^3G^2\log(2/\delta)}{2\mu^2}$, then with probability at least $1-\delta$, there exists a unit vector $\w[*]\in\R^{d'}$ such that $y\inner{\nphi(\w[],\x[])}{\w[*]}\ge\frac{\mu}{80d}>0$.
\end{example}

We now prove \cref{prop:linear,prop:ntk} with reference to \cite{taheri2024generalization}.
\begin{proof}[Proof of \cref{prop:linear}]
	For any fixed $\epsilon>0$, set $\alpha=2\log(1/\epsilon)/(\gamma\sqrt{m})$ and assume $m\ge4\log^2(1/\epsilon)$. Select weights $\wj[j]_{\epsilon}:=\alpha\vv^*, a_j=1$ for $j\in\left[\frac{m}{2}\right]$ and $\wj[j]_{\epsilon}:=-\alpha\vv^*, a_j=-1$ for $j\in\left\{\frac{m}{2}+1,\dots,m\right\}$.
	Then for any $(\x,y_i)$, we have
	\begin{align}
		y_i\Phi(\w[\epsilon],\x) &= \frac{y_i\sqrt{m}}{2}\left(\sigma\left(\alpha\inner{\vv^*}{\x}\right)-\sigma\left(-\alpha\inner{\vv^*}{\x}\right)\right) 
		= y_i\sqrt{m}\sigma\left(\alpha\inner{\vv^*}{\x}\right) \nonumber\\
		&\ge \sqrt{m} \sigma(\alpha\gamma)
		\ge \frac{\sqrt{m}}{2} \alpha \gamma 
		= \log(1/\epsilon), \nonumber
	\end{align}
	where the second equality uses the fact that tanh is odd, the first inequality follows the monotonicity of tanh and data separability condition, and the last inequality follows since $\alpha\gamma\le1$ and $\sigma(u)\ge u/2$ for all $u\in[0,1]$.
	Thus, $F(\w[\epsilon];\z)\le e^{-y_i\Phi(\w[\epsilon],\x)}\le e^{-\log(1/\epsilon)}=\epsilon$ since $\log(1+e^{-u})\le e^{-u}$, while $g(\epsilon)=\norm{\w[\epsilon]}=\alpha\sqrt{m}=2\log(1/\epsilon)/\gamma$. 
\end{proof}
\begin{proof}[Proof of \cref{prop:ntk}]
	Take $\w[\epsilon]=\w[1]+\frac{\w[*]}{\gamma}\left(2C+\log(1/\epsilon)\right)$. Since $\norm{\w[*]}=1$, we derive that $\norm{\w[\epsilon]-\w[1]}=\frac{2C+\log(1/\epsilon)}{\gamma}$.
	By Taylor's expansion, there exists $\w[]\in[\w[\epsilon],\w[1]]$ such that
	\begin{align}
		y_i\Phi\left(\w[\epsilon],\x\right) &=
		y_i\Phi\left(\w[1],\x\right)
		+ y_i\inner{\nphi\left(\w[1],\x\right)}{\w[\epsilon]-\w[1]}
		+ \frac{1}{2}y_i\inner{\w[\epsilon]-\w[1]}{\hphi\left(\w[],\x\right)\left(\w[\epsilon]-\w[1]\right)} \nonumber\\
		&\geq
		-\abs{\Phi\left(\w[1],\x\right)}
		+ \frac{y_i\left(2C+\log(1/\epsilon)\right)}{\gamma}\inner{\nphi\left(\w[1],\x\right)}{\w[*]}
		- \frac{\left(2C+\log(1/\epsilon)\right)^2}{2\gamma^2}\norm{\hphi\left(\w[],\x\right)} \nonumber\\
		&\geq 
		-C + 2C + \log(1/\epsilon) - \frac{R_2\left(2C+\log(1/\epsilon)\right)^2}{2\gamma^2\sqrt{m}}
		\geq \log(1/\epsilon), \nonumber
	\end{align}
	where we apply \cref{lem:s-g-h} in the second inequality and the condition on $m$ on the last one. Thus, $F(\w[\epsilon];\z)\le\epsilon$ since $\log(1+e^{-u})\le e^{-u}$.
\end{proof}

\section{Properties of Shallow Neural Networks}\label{sc:pf-properties}

The proofs of \cref{lem:s-g-h,lem:s-weakly-properties} follow a similar reasoning in \cite{taheri2024generalization}.
Specifically, \cref{lem:s-g-h} is identical to Lemma D.1 in the literature \cite{taheri2024generalization} and its proof is omitted. Since \cref{lem:s-weakly-properties} covers both per-sample gradients and full-batch gradients, we include the proof here for completeness.

\begin{proof}[Proof of \cref{lem:s-weakly-properties}]
	For notational simplicity, denote $\nphi=\nphi\left(\w[],\x[]\right)$ and $\hphi=\hphi\left(\w[],\x[]\right)$. For any $\w[]$ and $\z[]=(\x[],y)$, the gradient of $F(\w[];\z[])$ is 
	\begin{align}
		\nf{\w[]}{\z[]} &= f'\left(y\Phi(\w[],\x[])\right)y\nphi. \nonumber
	\end{align}
	Recalling that $y\in\{\pm1\}$ and applying \cref{lem:s-g-h}, we have
	\begin{align}
		\norm{\nf{\w[]}{\z[]}} &= \norm{f'\left(y\Phi(\w[],\x[])\right)y\nphi}
		= \abs{f'\left(y\Phi(\w[],\x[])\right)}\norm{\nphi}
		\le R_1\abs{f'\left(y\Phi(\w[],\x[])\right)}. \nonumber
	\end{align}
	Using that $\abs{f'(u)}\le\beta_1$ and $\abs{f'(u)}\le\beta_3f(u)$, we get, respectively,
	\begin{align}
		\norm{\nf{\w[]}{\z[]}} \le R_1{f'(y\nphi(\w[],\x[]))} \le \beta_1R_1 \quad \mbox{and} \quad
		\norm{\nf{\w[]}{\z[]}} \le R_1{f'(y\nphi(\w[],\x[]))} \le \beta_3R_1F(\w[];\z[]). \nonumber
	\end{align}
	Also, since $f''(u)\le\beta_2$, we obtain $\abs{f'(u)}^2\le 2\beta_2 f(u)$. Hence,
	\begin{align}
		\norm{\nf{\w[]}{\z[]}} \le R_1{f'(y\nphi(\w[],\x[]))} \le R_1\sqrt{2\beta_2F(\w[];\z[])}. \nonumber
	\end{align}
	For the full-batch gradients, we obtain $\norm{\nabla\F(\w[])} \le\frac{1}{n}\sum_{i=1}^{n}\norm{\nf{\w[]}{\z}} 
	\le \beta_1R_1$, $\norm{\nabla\F(\w[])}  \le \frac{\beta_3R_1}{n}\sum_{i=1}^{n}F(\w[];\z)
	= \beta_3R_1\F(\w[])$ and
	\begin{align} 
		\norm{\nabla\F(\w[])}^2 &\le 
		\norm{\frac{1}{n}\sum_{i=1}^{n}F(\w[];\z)}^2
		\le \frac{1}{n}\sum_{i=1}^{n}\norm{F(\w[];\z)}^2
		\le \frac{2\beta_2R_1^2}{n}\sum_{i=1}^{n}F(\w[];\z[])
		= 2\beta_2R_1^2\F(\w[]). \nonumber
	\end{align}
	
	For the Hessian, we derive
	\begin{align}\label{eq:s-h}
		\nfh{\w[]}{\z[]} &= f''\left(y\Phi(\w[],\x[])\right)y^2 \nphi\nphi\T
		+ f'(y\Phi(\w[],\x[]))y\hphi. 
	\end{align}
	Applying triangle inequality and \cref{lem:s-g-h}, we obtain
	\begin{align}
		\norm{\nfh{\w[]}{\z[]}} &\le \abs{f''\left(y\Phi(\w[],\x[])\right)} \norm{\nphi}^2
		+ \abs{f'(y\Phi(\w[],\x[]))}\norm{\hphi} \nonumber\\
		&\le R_1^2\abs{f''(y\Phi(\w[];\x[]))} + \frac{R_2}{\sqrt{m}}\abs{f'(y\Phi(\w[];\x[]))}
		\le R_1^2\beta_2+\frac{R_2\beta_1}{\sqrt{m}}, \nonumber
	\end{align}
	where in the last inequality we use that $\abs{f'(u)}\le\beta_1$ and $\abs{f''(u)}\le\beta_2$-smooth.
	
	Since $f$ is convex, $f''(\cdot)\ge0$, so the first term in \cref{eq:s-h} is positive semi-definite. For the second term, we obtain the lower bound
	\begin{align}
		\lambda_{\min}\left(\nfh{\w[]}{\z[]}\right) &\ge - \norm{f'(y\Phi(\w[],\x[]))\hphi}
		= - \abs{f'(y\Phi(\w[],\x[]))}\norm{\hphi}
		\ge - \frac{R_2\beta_1}{\sqrt{m}} \nonumber
	\end{align}
	or
	\begin{align}
		\lambda_{\min}\left(\nfh{\w[]}{\z[]}\right) &\ge - \frac{R_2\beta_3}{\sqrt{m}}F(\w[];\z[]). \nonumber
	\end{align}
	where we apply \cref{lem:s-weakly-properties,as:logistic}.
	The properties of the Hessian for the objective function $\F(\cdot)$ hold immediately since $\nfhf{\w[]}=\frac{1}{n}\sum_{i=1}^{n}\nfh{\w[]}{\z}$.
\end{proof}

\begin{proof}[Proof of \cref{lem:s-weakly-convex}]
	By Taylor's theorem, there exists an $\w[*]\in[\w[],\w[]']$ such that
	\begin{align}
		F(\w[]') &= F(\w[]) 
		+ \inner{\nabla F(\w[])}{\w[]'-\w[]}
		+ \frac{1}{2}\inner{\w[]'-\w[]}{\nabla^2 F(\w[*])\left(\w[]'-\w[]\right)} \nonumber\\
		&\ge F(\w[]) 
		+ \inner{\nabla F(\w[])}{\w[]'-\w[]}
		+ \frac{\norm{\w[]'-\w[]}^2}{2}\lambda_{\min}\left(\nabla^2 F(\w[*])\right). \nonumber
	\end{align}
	Applying \cref{lem:s-weakly-properties} yields that
	\begin{align}
		F(\w[]')
		&\ge F(\w[]) 
		+ \inner{\nabla F(\w[])}{\w[]'-\w[]}
		- \frac{C_2}{2\sqrt{m}}\norm{\w[]'-\w[]}^2, \nonumber
	\end{align}
	and
	\begin{align}
		F(\w[]') 
		&\ge F(\w[]) 
		+ \inner{\nabla F(\w[])}{\w[]'-\w[]}
		- \frac{C_2D^2}{2\sqrt{m}} F(\w[*]) \nonumber\\
		&\ge F(\w[]) 
		+ \inner{\nabla F(\w[])}{\w[]'-\w[]}
		- \frac{C_2D^2}{2\sqrt{m}} \max_{\vv\in[\w[],\w[]']} F(\vv). \qedhere \nonumber
	\end{align} 
\end{proof}

\begin{proof}[Proof of \cref{lem:s-max}]
	Let $\vv_*=\arg\max_{\vv\in[\w[],\w[]']} F(\vv)$. If $\vv_*=\w[]$ or $\vv_*=\w[]'$, then the result is trivial. Otherwise, $\vv_*$ lies in the interior of the segment. By the first-order optimal condition, we have $\inner{\nabla F(\vv_*)}{\w[]-\w[]'}=0$.
	
	Let $\vv_*=\alpha_*\w[]+(1-\alpha_*)\w[]', \alpha_*\in(0,1)$. By Taylor's theorem, there exists an $\w[*]\in[\w[],\vv_*]$ such that 
	\begin{align}
		F(\w[]) &= F(\vv_*)
		+ \inner{\nabla F(\vv_*)}{\w[]-\vv_*}
		+ \frac{1}{2}\inner{\vv_*-\w[]}{\nabla^2 F(\w[*])\left(\vv_*-\w[]\right)} \nonumber\\
		&= F(\vv_*)
		+ (1-\alpha_*)\inner{\nabla F(\vv_*)}{\w[]-\w[]'}
		+ \frac{1}{2}\inner{\vv_*-\w[]}{\nabla^2 F(\w[*])\left(\vv_*-\w[]\right)} \nonumber\\
		&= F(\vv_*)
		+ \frac{1}{2}\inner{\vv_*-\w[]}{\nabla^2 F(\w[*])\left(\vv_*-\w[]\right)}
		\ge F(\vv_*)
		+ \frac{\norm{\vv_*-\w[]}^2}{2}\lambda_{\min}\left(\nabla^2 F(\w[*])\right) \nonumber\\
		&\ge F(\vv_*)
		- \frac{C_2(1-\alpha_*)^2\norm{\w[]-\w[]'}^2}{2\sqrt{m}}F(\w[*])
		\ge 
		\left(1 - \frac{C_2D^2}{2\sqrt{m}}\right)F(\vv_*)
		\ge \frac{3}{4}F(\vv_*), \nonumber
	\end{align} 
	where the second inequality uses \cref{lem:s-weakly-properties}, the third inequality uses $F(\w[*])\le F(\vv_*)$ and the last step uses $\sqrt{m}\ge2C_2D^2$. Using the same deduction, we can also derive that $F(\w[]') \ge \frac{3}{4}F(\vv_*)$.
	Combining the above inequalities, we get the desired result.
\end{proof}

The following GD-expansiveness lemma is adapted from \cite{taheri2024generalization}.
\begin{lemma}[Lemma 5.2, \cite{taheri2024generalization}] \label{lem:expansive}
	Let \crefassump hold. Assume $F(\w[])$ satisfies $\norm{\nabla^2 F(\w[])}\le\theta+\kappa$ and $\lambda_{\min}\left(\nabla^2F(\w[])\right)\ge-\kappa F(\w[])$ for $\theta,\kappa\ge0$. For any $\w[],\w[]'\in\R^{d'}$ and $\w[\alpha]=\alpha\w[]+(1-\alpha)\w[]'$, 
	\begin{align}
		& \norm{\w[]-\eta\nabla F(\w[])-\w[]'+\eta\nabla F(\w[]')} \le \max_{\alpha\in[0,1]}H(\w[\alpha])\norm{\w[]-\w[]'}, \nonumber\\
		& H(\w[]) := \max\left\{1+\eta\kappa F(\w[]),\eta(\theta+\kappa)\right\}. \nonumber	
	\end{align}
\end{lemma}

\section{Omitted Proofs in the Full-batch Setting}\label{sc:pf-f}
\begin{proof}[Proof of \cref{thm:fs-conv}]
	We prove by induction that for all $s\in[T]$, 
	\begin{align}\label{eq:fs-hypo}
		\norm{\w[s]-\w[]}\le2\norm{\w[1]-\w[]}.
	\end{align}
	\textbf{Case $s=1$.} The inequality holds trivially.
	
	\noindent\textbf{Case $s=k+1$.} For $k\in[T]$, assume that \cref{eq:fs-hypo} holds for any $t\in[k]$. Then \cref{lem:s-weakly-convex,lem:s-max} hold for $\w[t]$ and $\w[]$ with $D=2\norm{\w[1]-\w[]}$ since $m\ge 64C_2^2\norm{\w[1]-\w[]}^4=4C_2^2D^4$.
	
	For any $t\in[k]$, we make the decomposition as
	\begin{align}
		\norm{\w[t+1]-\w[]}^2
		&= \norm{\w-\eta\nff{\tw}-\w[]}^2 \nonumber\\
		&= \norm{\w-\w[]}^2 
		-2\eta\inner{\nff{\tw}}{\w-\w[]} + \eta^2\norm{\nff{\tw}}^2 \nonumber\\
		&= \norm{\w-\w[]}^2
		+ 2\eta\inner{\nff{\w}-\nff{\tw}}{\w-\w[]} \nonumber\\
		&\quad
		- 2\eta\inner{\nff{\w}}{\w-\w[]}
		+ \eta^2\norm{\nff{\tw}}^2 \nonumber\\
		&\le \norm{\w-\w[]}^2
		+ 2\eta\rho L_F\norm{\w-\w[]} 
		\underbrace{- 2\eta\inner{\nff{\w}}{\w-\w[]}}_{\bm{I}}
		+ \underbrace{\eta^2\norm{\nff{\tw}}^2}_{\bm{II}}, \nonumber 
	\end{align}
	where we use Cauchy-Schwarz inequality and \cref{lem:s-weakly-properties} in the inequality.
	Applying \cref{lem:s-weakly-convex,lem:s-max}, we obtain that
	\begin{align}
		\bm{I} &\le 
		2\eta \F(\w[]) - 2\eta \F(\w)
		+ \frac{\eta C_2}{2\sqrt{m}}\max_{\vv\in[\w,\w[]]}\F(\vv)\norm{\w[t]-\w[]} \nonumber\\
		&\le 2\eta \F(\w[]) - 2\eta \F(\w)
		+ \frac{2\eta}{3}\max\left\{\F(\w[]),\F(\w)\right\} \nonumber\\
		&\le 
		\frac{8\eta}{3}\F(\w[])
		- \frac{4\eta}{3}\F(\w). \nonumber
	\end{align}
	For the second term, the triangle inequality and \Cref{lem:s-weakly-properties} give
	\begin{align}
		\bm{II}
		&= \eta^2\norm{\nff{\tw}-\nff{\w}+\nff{\w}} \nonumber\\
		&\le 2\eta^2\norm{\nff{\tw}-\nff{\w}}^2 
		+ 2\eta^2\norm{\nff{\w}}^2 \nonumber\\
		&\le 2\eta^2\rho^2L_F^2
		+ 2\eta^2C_1^2 \F(\w). \nonumber
	\end{align}
	Combining these inequalities with $\eta\le\min\left\{\frac{1}{6C_1^2},\frac{1}{4L_F}\right\}$ and $\rho\le\frac{2\norm{\w[1]-\w[]}}{T}$, we obtain
	\begin{align}
		\norm{\w[t+1]-\w[]}^2
		&\le \norm{\w-\w[]}^2
		+ \frac{8\eta}{3}\F(\w[])
		- \eta \F(\w)
		+ \frac{5\norm{\w[1]-\w[]}^2}{2T}. \nonumber
	\end{align}
	Summing up the above inequality over $t\in[k]$,
	\begin{align}\label{eq:fs-iteration}
		\norm{\w[k+1]-\w[]}^2
		&\le 
		\norm{\w[1]-\w[]}^2
		- \sum_{t=1}^{k}\eta \F(\w)
		+ \frac{8k}{3}\eta \F(\w[])
		+ \frac{5\norm{\w[1]-\w[]}^2}{2} \nonumber\\
		&\le 
		\frac{23}{6}\norm{\w[1]-\w[]}^2
		- \sum_{t=1}^{k}\eta \F(\w)
		\le 4\norm{\w[1]-\w[]}^2,
	\end{align}
	where we use \cref{eq:f-w} in the second inequality.
	
	Hence, \cref{eq:fs-hypo} holds for all $s\in[T]$.
	Based on the above argument, \eqref{eq:fs-iteration} is valid for all $k\in[T]$. Consequently, we derive that
	\begin{align}
		\sum_{t=1}^{k} \F(\w)
		&\le 
		\frac{4\norm{\w[1]-\w[]}^2}{\eta}. \qedhere \nonumber
	\end{align}
\end{proof}

\begin{proof}[Proof of \cref{thm:fs-stability}]
	Since $F(\w[];\z[])\ge0$ for all $\w[]$ and $\z[]$, we have
	\begin{align}
		\norm{\w[]-\w[1]}^2 \ge 8\eta T\F(\w[]) \ge 8\eta T\F[\Si](\w[]), \nonumber
	\end{align}
	which shows that $\wi$ and $\F[\Si]$ also satisfy the condition of \cref{thm:fs-conv}.
	Thus, by \cref{thm:fs-conv}, \cref{eq:fs-hypo} holds for $\w$ and $\wi, t\in[T]$.
	Consequently, \cref{lem:s-max} holds for $\w$ and $\wi$, since
	\begin{align}
		\norm{\w-\wi}^2 \le \left(\norm{\w-\w[]}+\norm{\wi-\w[]}\right)^2 \le 16\norm{\w[1]-\w[]}^2\le \frac{\sqrt{m}}{2C_2}, \nonumber
	\end{align}
	Using the update rule and the triangle inequality, we have
	\begin{align}
		\norm{\w[t+1]-\wi[t+1]}
		&= \norm{\w-\eta\nabla\F\left(\tw\right)-\wi+\eta\nabla\F[\Si]\left(\twi\right)} \nonumber\\
		&\le \norm{\w-\eta\nabla\F[\Si]\left(\w\right)-\wi+\eta\nabla\F[\Si]\left(\wi\right)} 
		+ \frac{\eta}{n}\norm{\nabla F\left(\w;\z\right)} \nonumber\\
		&\quad
		+ \eta\norm{\nabla\F\left(\w\right)-\nabla\F\left(\tw\right)} 
		+ \eta\norm{\nabla\F[\Si]\left(\wi\right)-\nabla\F[\Si]\left(\twi\right)} \nonumber\\
		&\le \max_{\vv\in\left[\w,\wi\right]}\left\{1+\frac{\eta C_2}{\sqrt{m}}\F[\Si](\vv),\eta L_F \right\}\norm{\w-\wi} 
		+ \frac{\eta C_1F(\w;\z)}{n}
		+ 2\eta\rho L_F \nonumber\\
		&\le \left(1+\frac{4\eta C_2}{3\sqrt{m}}\left(\F[\Si](\w)+ \F[\Si]\left(\wi\right)\right)\right)\norm{\w-\wi} 
		+ \frac{\eta C_1F(\w;\z)}{n}
		+ 2\eta\rho L_F \nonumber\\
		&= \left(1+\bi{t}\right)\norm{\w-\wi} 
		+ \Bi{t}, \nonumber
	\end{align}
	where we use \cref{lem:expansive,lem:s-weakly-properties} in the second inequality, \cref{lem:s-max} in the third one, and in the last equality we define $\bi{t}=\frac{4\eta C_2}{3\sqrt{m}}\left(\F[\Si](\w)+ \F[\Si]\left(\wi\right)\right)$, $\Bi{t}=\frac{\eta C_1}{n}F(\w;\z) + 2\eta\rho L_F$.
	
	Unrolling this recursion gives
	\begin{align}
		\norm{\w[t+1]-\wi[t+1]} &\le \sum_{s=1}^{t}\prod_{p=s+1}^{t}\left(1+\bi{p}\right)\Bi{s} 
		\le \sum_{s=1}^{t}\exp\left(\sum_{p=s+1}^{t}\bi{p}\right)\Bi{s} \nonumber\\
		&\le \exp\left(\sum_{s=1}^{t}\bi{s}\right)\sum_{s=1}^{t}\Bi{s} 
		\le \exp\left(\max_{j\in[n]}\sum_{s=1}^{t}\bi[j]{s}\right)\sum_{s=1}^{t}\Bi{s}. \nonumber
	\end{align}
	Now we bound the exponential factor:
	\begin{align}
		\max_{j\in[n]}\sum_{s=1}^{t}\bi[j]{s} &\le \frac{4\eta C_2}{3\sqrt{m}}\max_{i\in[n]}\left\{\sum_{s=1}^{t}\F[\Si](\w[s])+ \sum_{s=1}^{t}\F[\Si]\left(\wi[s]\right)\right\} \nonumber\\
		&\le \frac{4\eta C_2}{3\sqrt{m}}\max_{i\in[n]}\left\{\sum_{s=1}^{t}\F(\w[s])+\sum_{s=1}^{t}\F[\Si]\left(\wi[s]\right)\right\} 
		\le \frac{4}{3}, \nonumber
	\end{align}
	where we use $\sqrt{m}\ge\eta C_2\max_{i\in[n]}\left\{\sum_{t=1}^{T}\F(\w)+ \sum_{t=1}^{T}\F[\Si]\left(\wi\right)\right\}$ in the last inequality. 
	Therefore,
	\begin{align}
		\frac{1}{n}\sum_{i=1}^{n}\E\left[\norm{\w[t+1]-\wi[t+1]}\right] &\le \frac{1}{n}\sum_{i=1}^{n}\sum_{s=1}^{t}\frac{4\eta C_1}{n}\E\left[F(\w[s];\z)\right] + 8\eta\rho L_Ft =\frac{4\eta C_1}{n}\sum_{s=1}^{t}\E\left[\F(\w)\right] + 8\eta\rho L_Ft. \qedhere \nonumber
	\end{align}
\end{proof}

\begin{proof}[Proof of \cref{thm:ff-gen}]
	Under the conditions of \cref{thm:ff-gen}, we can verify that \cref{eq:fs-1} also holds. Averaging \cref{eq:fs-1} over $t=1,\dots,T$ and rearranging, we obtain
	\begin{align}
		\E\left[\F[\D]\left(\w\right)-\F(\w)\right]
		&\le \frac{2e\eta C_1^2}{n}\sum_{s=1}^{t-1}\E\left[\F(\w[s])\right] + 4e\eta\rho C_1L_Ft \nonumber\\
		&\le 
		\frac{8e C_1\norm{\w[1]-\w[]}^2}{n} 
		+ 4e\eta\rho L_FT, 
	\end{align}
	\begin{align}
		\frac{1}{T}\sum_{t=1}^{T}\E\left[\F[\D](\w)\right]
		&\le 
		\frac{1}{T}\sum_{t=1}^{T}\E\left[\F(\w)\right]
		+ \frac{8e C_1^2\norm{\w[1]-\w[]}^2}{n}
		+ 4e\eta\rho C_1L_FT \nonumber\\
		&\le
		\left(1+\frac{2e\eta C_1^2T}{n}\right)\frac{4\norm{\w[1]-\w[]}^2}{\eta T}
		+ 4e\eta\rho C_1L_FT \nonumber\\
		&\le
		\cO\left(\left(1+\frac{\sqrt{T}}{n}\right)\frac{\norm{\w[1]-\w[]}^2}{\eta T}
		+ \frac{\norm{\w[1]-\w[]}}{\sqrt{T}}\right), \nonumber
	\end{align}
	where the second line uses \Cref{thm:fs-conv}, and the last step follows from the step‑size and perturbation bounds $\eta\le\min\left\{\frac{1}{C_1^2},\frac{1}{C_1L_F\sqrt{T}}\right\}$ and $\rho\le\frac{2\norm{\w[1]-\w[]}}{T}$.
	
	If, additionally, \cref{as:f-nn-realizability} holds, we may take $\w[]=\w[\epsilon]$ with $\F(\w[\epsilon])\le\epsilon$, yielding 
	\begin{align}
		\frac{1}{T}\sum_{t=1}^{T}\E\left[\F[\D](\w)\right]
		&\le
		\cO\left(\left(\frac{1}{T}+\frac{1}{n\sqrt{T}}\right)\frac{g(\epsilon)^2}{\eta} + \frac{g(\epsilon)}{\sqrt{T}}\right). \qedhere\nonumber
	\end{align}
\end{proof}

\section{Proof in the Stochastic Setting}\label{sc:pf-s}
\subsection{Proof of \cref{thm:s-gen}}
\begin{proposition}[Convergence Bounds]\label{thm:s-conv}
	Assume the same conditions as in \cref{thm:s-gen}. 
	We have
	\begin{align}
		&\sum_{s=1}^{t} \E\left[\F(\w[s])\right]\le \frac{4\norm{\w[1]-\w[]}^2}{\eta},\ \forall t\in[T]. \nonumber
	\end{align}
	Additionally, we have $\norm{\w-\w[]}\le2\norm{\w[1]-\w[]}$ holds for any $t\in[T]$.
\end{proposition}
\begin{proof}
	We prove by induction that for all $s\in[T]$,
	\begin{align}\label{eq:s-hypo}
		\norm{\w[s]-\w[]}\le2\norm{\w[1]-\w[]}.
	\end{align}
	\textbf{Case $s=1$.} The inequality holds trivially.
	
	\noindent\textbf{Case $s=k+1$.} For $k\in[T]$, assume that \cref{eq:s-hypo} holds for any $t\in[k]$. Then, \cref{lem:s-weakly-convex,lem:s-max} hold for $\w[t]$ and $\w[]$ with $D=2\norm{\w[1]-\w[]}$ since $m\ge 16C_2^2\norm{\w[1]-\w[]}^4=4C_2^2D^2$.
	
	For any $t\in[k]$, we make the decomposition as
	\begin{align}
		\norm{\w[t+1]-\w[]}^2
		&= \norm{\w-\nf{\tw}{\z[i_t]}-\w[]}^2 \nonumber\\
		&= \norm{\w-\w[]}^2 
		-2\eta\inner{\nf{\tw}{\z[i_t]}}{\w-\w[]} + \eta^2\norm{\nf{\tw}{\z[i_t]}}^2 \nonumber\\
		&= \norm{\w-\w[]}^2
		+ 2\eta\inner{\nf{\w}{\z[i_t]}-\nf{\tw}{\z[i_t]}}{\w-\w[]} \nonumber\\
		&\quad
		- 2\eta\inner{\nf{\w}{\z[i_t]}}{\w-\w[]}
		+ \eta^2\norm{\nf{\tw}{\z[i_t]}}^2 \nonumber\\
		&\le \norm{\w-\w[]}^2
		+ 2\eta\rho L_F\norm{\w-\w[]} 
		\underbrace{- 2\eta\inner{\nf{\w}{\z[i_t]}}{\w-\w[]}}_{\bm{I}}
		+ \underbrace{\eta^2\norm{\nf{\tw}{\z[i_t]}}^2}_{\bm{II}}, \nonumber 
	\end{align}
	where we use the Cauchy-Schwarz inequality and \cref{lem:s-weakly-properties} in the inequality.
	Applying \cref{lem:s-weakly-convex,lem:s-max}, we obtain that
	\begin{align}
		\bm{I} &\le 
		2\eta F(\w[];\z[i_t]) - 2\eta F(\w;\z[i_t])
		+ \frac{\eta}{2}\max_{\vv\in[\w,\w[]]}F(\vv;\z[i_t]) \nonumber\\
		&\le 2\eta F(\w[];\z[i_t]) - 2\eta F(\w;\z[i_t])
		+ \frac{2\eta}{3}\max\left\{F(\w[];\z[i_t]),F(\w;\z[i_t])\right\} \nonumber\\
		&\le 
		\frac{8\eta}{3}F(\w[];\z[i_t])
		- \frac{4\eta}{3}F(\w;\z[i_t]). \nonumber
	\end{align}
	For the second term, the triangle inequality and \cref{lem:s-weakly-properties} give
	\begin{align}
		\bm{II}
		&= \eta^2\norm{\nf{\tw}{\z[i_t]}-\nf{\w}{\z[i_t]}+\nf{\w}{\z[i_t]}} \nonumber\\
		&\le 2\eta^2\norm{\nf{\tw}{\z[i_t]}-\nf{\w}{\z[i_t]}}^2 
		+ 2\eta^2\norm{\nf{\w}{\z[i_t]}}^2 \nonumber\\
		&\le 2\eta^2\rho^2L_F^2
		+ 2\eta^2C_1^2 F(\w;\z[i_t]). \nonumber
	\end{align}
	Combining these inequalities and using $\eta\le\min\left\{\frac{1}{6C_1^2},\frac{1}{4L_F}\right\}$ and $\rho\le\frac{2\norm{\w[1]-\w[]}}{T}$, we obtain
	\begin{align} 
		\norm{\w[t+1]-\w[]}^2
		&\le \norm{\w-\w[]}^2
		+ \frac{8\eta}{3}F(\w[];\z[i_t])
		- \eta F(\w;\z[i_t])
		+ \frac{5\norm{\w[1]-\w[]}^2}{2T}. \nonumber
	\end{align}
	Summing up the above inequality over $t\in[k]$ and using \cref{eq:s-w}, we get
	\begin{align}\label{eq:s-iteration}
		\norm{\w[k+1]-\w[]}^2
		&\le 
		\norm{\w[1]-\w[]}^2
		- \sum_{t=1}^{k}\eta F(\w;\z[i_t])
		+ \frac{8k}{3}\eta F(\w[];\z[i_t])
		+ \frac{5\norm{\w[1]-\w[]}^2}{2} \nonumber\\
		&\le 
		\frac{4}{3}\norm{\w[1]-\w[]}^2
		- \sum_{t=1}^{k}\eta F(\w;\z[i_t])
		+ \frac{5\norm{\w[1]-\w[]}^2}{2}
		\le 4\norm{\w[1]-\w[]}^2. 
	\end{align}
	Hence, the induction is complete. Based on the above argument, \cref{eq:s-iteration} holds for all $k\in[T]$.
	Taking expectation \cref{eq:s-iteration}, we obtain
	\begin{align}
		\sum_{t=1}^{k} \E\left[\F(\w)\right]
		&\le
		\frac{4\norm{\w[1]-\w[]}^2}{\eta}. \qedhere \nonumber
	\end{align}
\end{proof}

\begin{proposition}\label{thm:s-stability}
	Assume the same conditions as in \cref{thm:s-gen}. 
	Let $S'=\{\z[1]',\dots,\z[n]'\}$ be drawn independently from $\D$. For any $i\in[n]$, let $\Si=\left\{\z[1],\dots,\z[i-1],\z[i]',\z[i+1],\dots,\z[n]\right\}$ be the training set obtained from $S$ by replacing $\z$ with $\z'$, and let $\left\{\wi\right\}_{t=1}^{T}$ be produced by the stochastic variant of \cref{alg} on $\Si$. Then for any $t\in[T]$,
	\begin{align}
		\max_{t\in[T]}\frac{1}{n}\sum_{i=1}^{n}\E\left[\norm{\w-\wi}\right]
		&\leq
		\frac{4\eta C_1}{n}\sum_{t=1}^{T}\E\left[\F(\w)\right]
		+ 4\eta\rho L_FT. \nonumber
	\end{align}
\end{proposition}
\begin{proof}
	By the update rule,
	\begin{align}
		\norm{\w[t+1]-\wi[t+1]}
		&= 
		\norm{\w-\eta\nf{\tw}{\z[i_t]}-\wi+\eta\nf{\twi}{\z[i_t]'}} \nonumber\\
		&\le \norm{\w-\wi} + \eta\norm{\nf{\tw}{\z[i_t]}-\nf{\twi}{\z[i_t]'}}. \nonumber
	\end{align}
	Decompose the gradient term as
	\begin{align}
		\norm{\nf{\tw}{\z[i_t]}-\nf{\twi}{\z[i_t]'}}
		&\le 
		\norm{\nf{\tw}{\z[i_t]}-\nf{\w}{\z[i_t]}}
		+ \norm{\nf{\w}{\z[i_t]}-\nf{\wi}{\z[i_t]'}} \nonumber\\
		&\quad
		+ \norm{\nf{\wi}{\z[i_t]'}-\nf{\twi}{\z[i_t]'}} \nonumber\\
		&\le 
		\norm{\nf{\w}{\z[i_t]}-\nf{\wi}{\z[i_t]'}}
		+ 2\rho L_F, \nonumber
	\end{align}
	where we use \cref{lem:s-weakly-properties} in the last inequality.
	Inserting this back and taking expectation implies
	\begin{align}
		\E\left[\norm{\w[t+1]-\wi[t+1]}\right]
		&=
		\E\left[\norm{\w-\wi}\right]
		+ \eta\E\left[\norm{\nf{\w}{\z[i_t]}-\nf{\wi}{\z[i_t]'}}\right] 
		+ 2\eta\rho L_F. \nonumber
	\end{align}
	Noting the distribution of $i_t$, we have
	\begin{align}
		&\ \E\left[\norm{\nf{\w}{\z[i_t]}-\nf{\wi}{\z[i_t]'}}\right] 
		= \E\left[\Et\left[\norm{\nf{\w}{\z[i_t]}-\nf{\wi}{\z[i_t]'}}\right]\right] \nonumber\\
		=&\  
		\E\left[\frac{1}{n}\sum_{j\ne i}\norm{\nf{\w}{\z[j]}-\nf{\wi}{\z[j]}}
		+ \frac{1}{n}\norm{\nf{\w}{\z}-\nf{\wi}{\z'}}\right] \nonumber\\
		\le&\ 
		L_F\E\left[\norm{\w-\wi}\right] 
		+ \frac{1}{n}\E\left[\norm{\nf{\w}{\z}}+\norm{\nf{\wi}{\z'}}\right] \nonumber\\
		\le&\ 
		L_F\E\left[\norm{\w-\wi}\right] 
		+ \frac{2}{n}\E\left[\norm{\nf{\w}{\z}}\right]. \nonumber
	\end{align}
	where we apply \cref{lem:s-weakly-properties} in the first inequality and note that $\E\left[\norm{\nf{\w}{\z}}\right]=\E\left[\norm{\nf{\wi}{\z'}}\right]$ in the last one since the symmetry between $\z$ and $\z'$ .
	
	Combining the above inequalities, we have
	\begin{align}
		\E\left[\norm{\w[t+1]-\wi[t+1]}\right]
		&\leq
		\left(1+\eta L_F\right)\E\left[\norm{\w-\wi}\right]
		+ \frac{2\eta}{n}\E\left[\norm{\nf{\w}{\z}}\right]
		+ 2\eta\rho L_F. \nonumber
	\end{align}
	Unrolling and using $\w[1]=\wi[1]$ leads to
	\begin{align}
		\E\left[\norm{\w[k+1]-\wi[k+1]}\right]
		&\leq
		\eta L_F\sum_{t=1}^{k}\E\left[\norm{\w-\wi}\right]
		+ \frac{2\eta}{n}\sum_{t=1}^{k}\E\left[\norm{\nf{\w}{\z}}\right]
		+ 2k\eta\rho L_F \nonumber\\
		&\leq
		\frac{1}{2}\max_{t\in[T]}\E\left[\norm{\w-\wi}\right]
		+ \frac{2\eta}{n}\sum_{t=1}^{k}\E\left[\norm{\nf{\w}{\z}}\right]
		+ 2k\eta\rho L_F, \nonumber
	\end{align}
	where we use $\eta\le\frac{1}{2L_FT}$. 
	Since the above inequality holds for any $k\in[T]$, we obtain
	\begin{align}
		\max_{t\in[T]}\E\left[\norm{\w-\wi}\right]
		&\leq \frac{4\eta}{n}\sum_{t=1}^{T}\E\left[\norm{\nf{\w}{\z}}\right]
		+ 4\eta\rho L_F 
		\leq 
		\frac{4\eta C_1}{n}\sum_{t=1}^{T}\E\left[F(\w;\z)\right]
		+ 4\eta\rho L_FT. \nonumber
	\end{align}
	Summing up the above inequality over $i\in[n]$, we obtain the results
	\begin{align}
		\max_{t\in[T]}\frac{1}{n}\sum_{i=1}^{n}\E\left[\norm{\w-\wi}\right]
		&\leq
		\frac{4\eta C_1}{n}\sum_{t=1}^{T}\E\left[\F(\w)\right]
		+ 4\eta\rho L_FT. \qedhere \nonumber
	\end{align}
\end{proof}

\begin{proof}[Proof of \cref{thm:s-gen}]
	We have
	\begin{align}
		\E\left[\F[\D](\w[T])-\F(\w[T])\right]
		&= 
		\frac{1}{n}\sum_{i=1}^{n}\E\left[F\left(\w[T];\z[]\right)-F\left(\w[T];\z\right)\right] \nonumber\\
		&= 
		\frac{1}{n}\sum_{i=1}^{n}\E\left[F(\w[T];\z[])-F\left(\wi[T];\z[]\right)\right]
		+ 
		\frac{1}{n}\sum_{i=1}^{n}\E\left[F\left(\wi[T];\z\right)-F\left(\w[T];\z\right)\right] \nonumber\\
		&\le 
		\frac{C_1}{n}\sum_{i=1}^{n}\E\left[\norm{\wi[T]-\w[T]}\right], \nonumber
	\end{align}
	where we apply \cref{lem:s-weakly-properties} in the last inequality. 
	Combined with \cref{thm:s-stability,thm:s-conv}, we obtain
	\begin{align}
		\E\left[\F[\D](\w[T])-\F(\w[T])\right] 
		&\leq
		\frac{32\eta C_1^2}{3n}\sum_{t=1}^{T}\E\left[\F(\w)\right]
		+ \frac{32\eta\rho L_FC_1T}{3}  \nonumber\\
		&\leq
		\frac{128 C_1^2}{3n} \norm{\w[1]-\w[]}^2
		+ \frac{32\eta\rho L_FC_1T}{3} \nonumber\\
		&\leq
		O\left(\frac{\norm{\w[1]-\w[]}^2}{n} 
		+ \rho \right), \nonumber
	\end{align}
	where we use $\eta\le\frac{1}{2L_FT}$ in the last inequality.
	
	Finally, if \cref{as:nn-realizability} holds with sufficiently small $\epsilon\le1/T$, we may take $\w[]$ as $\w[\epsilon]$, for which $\F(\w[\epsilon])\le\epsilon$ and $\norm{\w[1]-\w[\epsilon]}\le g(\epsilon)$. The step size $\eta \le \frac{\norm{\w[1]-\w[\epsilon]}^2}{8}$ then implies 
	\begin{align}
		8\eta 	\sum_{t=1}^{T}F(\w[\epsilon];\z[i_t]) \le T \epsilon \norm{\w[1]-\w[\epsilon]}^2  \le\norm{\w[1]-\w[\epsilon]}^2, \nonumber
	\end{align}
	so \cref{eq:s-w} is satisfied. Substituting these estimates into the previous bound yields
	\begin{align}
		\E\left[\F[\D]\left(\bar{\bm{w}}_T\right)\right]
		&\le
		O\left(\frac{g(\epsilon)^2}{n}
		+ \frac{g(\epsilon)}{T}\right). \qedhere \nonumber
	\end{align}
\end{proof}

\subsection{Proof of \cref{thm:ss-gen}}\label{sc:pf-ss}
\begin{proposition}\label{thm:ss-on-ave}
	Assume the same conditions as in \cref{thm:s-gen}. Let $S'=\{\z[1]',\dots,\z[n]'\}$ be drawn independently from $\D$. For any $i\in[n]$, let $\Si=\left\{\z[1],\dots,\z[i-1],\z[i]',\z[i+1],\dots,\z[n]\right\}$ be the set formed from $S$ by replacing the $i$-th element with $\z[i]'$ and $\left\{\wi\right\}_{t=1}^{T}$ be produced by the stochastic variant of \cref{alg} with training set $\Si$. Then we have for any $t\in[T]$,
	\begin{align}
		\E\left[\F[\D](\w)-\F(\w)\right] 
		\le 
		\frac{1+L_F}{2n}\sum_{i=1}^{n}\E\left[\norm{\wi-\w}^2\right]
		+ \frac{C_1^2}{2}\E\left[\F(\w)\right], \nonumber
	\end{align}
\end{proposition}
\begin{proof}
	Due to the symmetry of $S$ and $S'$, we have
	\begin{align}
		& \E\left[\F[\D](\w)-\F(\w)\right] \nonumber\\
		&= 
		\E\left[\frac{1}{n}\sum_{i=1}^{n}\F[\D](\wi)-\F(\w)\right]
		= \E\left[\frac{1}{n}\sum_{i=1}^{n}\left(F(\wi;\z)-F(\w;\z)\right)\right] \nonumber\\
		&\le 
		\frac{1}{n}\sum_{i=1}^{n}\E\left[\inner{\wi-\w}{\nf{\w}{\z}}+\frac{L_F}{2}\norm{\wi-\w}^2\right] \nonumber\\
		&\le 
		\frac{1}{n}\sum_{i=1}^{n}\E\left[
		\frac{1}{2}\norm{\wi-\w}^2
		+ \frac{1}{2}\norm{\nf{\w}{\z}}^2
		+ \frac{L_F}{2}\norm{\wi-\w}^2
		\right] \nonumber\\
		&\le 
		\frac{1+L_F}{2n}\sum_{i=1}^{n}\E\left[\norm{\wi-\w}^2\right]
		+ \frac{C_1^2}{2}\E\left[\F(\w)\right], \nonumber
	\end{align}
	where the first inequality uses \cref{lem:s-weakly-properties} in the second applies Young's inequality and the last again uses \cref{lem:s-weakly-properties}.
\end{proof}

\begin{proposition}\label{thm:ss-stability}
	Assume the same conditions as in \cref{thm:ss-on-ave}. 
	We have for any $t\in[T]$,
	\begin{align}
		\frac{1}{n}\sum_{i=1}^{n}\E \left[\norm{\w[t]-\wi[t]}^2\right]
		&\le \frac{36\eta^2C_1^2}{n}\left(1+\frac{2T}{n}\right)\sum_{t=1}^{T}\E\left[F(\w;\z)\right]
		+ 90\eta^2\rho ^2L_F^2T^2. \nonumber
	\end{align}
\end{proposition}
\begin{proof}
	By \cref{alg}, we have
	\begin{align}
		\norm{\w[t+1]-\wi[t+1]}^2
		&= 
		\norm{\w-\eta\nf{\tw}{\z[i_t]}-\wi+\eta\nf{\twi}{\z[i_t]'}}^2 \nonumber\\
		&= \norm{\w-\wi}^2 + \eta^2\norm{\nf{\tw}{\z[i_t]}-\nf{\twi}{\z[i_t]'}}^2 \nonumber\\
		&\quad
		-2\eta\inner{\w-\wi}{\nf{\tw}{\z[i_t]}-\nf{\twi}{\z[i_t]'}}. \nonumber
	\end{align}
	Applying norm inequality, we obtain
	\begin{align}
		\norm{\nf{\tw}{\z[i_t]}-\nf{\twi}{\z[i_t]'}}^2
		&\le 
		3\norm{\nf{\tw}{\z[i_t]}-\nf{\w}{\z[i_t]}}^2
		+ 3\norm{\nf{\w}{\z[i_t]}-\nf{\wi}{\z[i_t]'}}^2 \nonumber\\
		&\quad
		+ 3\norm{\nf{\wi}{\z[i_t]'}-\nf{\twi}{\z[i_t]'}}^2 \nonumber\\
		&\le 
		3\norm{\nf{\w}{\z[i_t]}-\nf{\wi}{\z[i_t]'}}^2
		+ 6\rho^2L_F^2, \nonumber
	\end{align}
	where we use \cref{lem:s-weakly-properties} in the last inequality. 
	Make a decomposition of the cross term as
	\begin{align}
		&\ -2\eta\inner{\w-\wi}{\nf{\tw}{\z[i_t]}-\nf{\twi}{\z[i_t]'}} \nonumber\\
		=&\ 
		-2\eta\inner{\w-\wi}{\nf{\tw}{\z[i_t]}-\nf{\w}{\z[i_t]}}
		- 2\eta\inner{\w-\wi}{\nf{\w}{\z[i_t]}-\nf{\wi}{\z[i_t]'}} \nonumber\\
		&\quad
		- 2\eta\inner{\w-\wi}{\nf{\wi}{\z[i_t]'}-\nf{\twi}{\z[i_t]'}} \nonumber\\
		\le &\ 
		4\eta\rho L_F\norm{\w-\wi} 
		- 2\eta\inner{\w-\wi}{\nf{\w}{\z[i_t]}-\nf{\wi}{\z[i_t]'}}, \nonumber
	\end{align}
	where we use Cauchy-Schwarz inequality and \cref{lem:s-weakly-properties} in the inequality.
	Combining the above inequalities, we obtain
	\begin{align}
		\norm{\w[t+1]-\wi[t+1]}^2
		&\le \norm{\w-\wi}^2 
		+ 3\eta^2\norm{\nf{\w}{\z[i_t]}-\nf{\wi}{\z[i_t]'}}^2
		+ 6\eta^2\rho^2L_F^2 \nonumber\\
		&\quad
		+ 4\eta\rho L_F\norm{\w-\wi} 
		- 2\eta\inner{\w-\wi}{\nf{\w}{\z[i_t]}-\nf{\wi}{\z[i_t]'}}, \nonumber
	\end{align}
	Taking a conditional expectation implies
	\begin{align}
		\Et\left[\norm{\w[t+1]-\wi[t+1]}^2\right]
		&=
		\norm{\w-\wi}^2
		+ 3\eta^2\Et\left[\norm{\nf{\w}{\z[i_t]}-\nf{\wi}{\z[i_t]'}}^2\right] 
		+ 6\eta^2\rho^2L_F^2 \nonumber\\
		&\quad
		+ 4\eta\rho L_F\norm{\w-\wi}
		- 2\eta\inner{\w-\wi}{\nabla\F(\w)-\nabla\F[\Si]\left(\wi\right)}. \nonumber
	\end{align}
	By \cref{lem:s-weakly-convex}, we have $\inner{\nabla F(\w[];\z[])-\nabla F(\w[]';\z[])}{\w[]-\w[]'} \ge 
	- \frac{C_2}{\sqrt{m}} \norm{\w[]-\w[]'}^2$ for any $\z[]$ and $\w[],\w[]'\in\R^{d'}$, and then	 
	\begin{align}
		&\ \inner{\w-\wi}{\nabla\F(\w)-\nabla\F[\Si]\left(\wi\right)} \nonumber\\
		=&\ 
		\inner{\w-\wi}{\frac{1}{n}\sum_{j=1}^{n}\nf{\w}{\z[j]}-\frac{1}{n}\sum_{j=1}^{n}\nf{\wi}{\z[j]'}} \nonumber\\
		=&\ 
		\frac{1}{n}\sum_{j\ne i}\inner{\w-\wi}{\nf{\w}{\z[j]}-\nf{\wi}{\z[j]}}
		+ \frac{1}{n}\inner{\w-\wi}{\nf{\w}{\z}-\nf{\wi}{\z'}} \nonumber\\
		\geq&\ 
		-\frac{(n-1)C_2}{\sqrt{m}n}\norm{\w-\wi}^2
		- \frac{1}{n}\norm{\w-\wi}\norm{\nf{\w}{\z}-\nf{\wi}{\z'}}. \nonumber
	\end{align}
	Thus, we obtain
	\begin{align}
		\E\left[\norm{\w[t+1]-\wi[t+1]}^2\right]
		&\leq
		\left(1+\frac{2\eta C_2}{\sqrt{m}}\right)\E\left[\norm{\w-\wi}^2\right] 
		+ 6\eta^2\rho^2L_F^2
		+ 4\eta\rho L_F\E\left[\norm{\w-\wi}\right] \nonumber\\
		&\quad 
		+ 3\eta^2\E\left[\norm{\nf{\w}{\z[i_t]}-\nf{\wi}{\z[i_t]'}}^2\right] \nonumber\\
		&\quad 
		+ \frac{2\eta}{n}\E\left[\norm{\w-\wi}\norm{\nf{\w}{\z}-\nf{\wi}{\z'}}\right]. \nonumber
	\end{align}
	Using Cauchy-Schwarz inequality and $(a+b)^2\le2\left(a^2+b^2\right)$, we obtain
	\begin{align}
		&\ 
		\E\left[\norm{\w-\wi}\left(2\rho L_F+\frac{1}{n}\norm{\nf{\w}{\z}-\nf{\wi}{\z'}}\right)\right] \nonumber\\
		\leq&\ 
		\left(\E\left[\norm{\w-\wi}^2\right]\right)^{1/2} \left(\E\left[\left(2\rho L_F+\frac{1}{n}\norm{\nf{\w}{\z}-\nf{\wi}{\z'}}\right)^2\right]\right)^{1/2} \nonumber\\
		\leq&\ 
		\sqrt{2}\left(\E\left[\norm{\w-\wi}^2\right]\right)^{1/2} \left(4\rho^2L_F^2+\frac{1}{n^2}\E\left[\norm{\nf{\w}{\z}-\nf{\wi}{\z'}}^2\right]\right)^{1/2} \nonumber\\
		\leq&\ 
		\sqrt{2}\left(\E\left[\norm{\w-\wi}^2\right]\right)^{1/2} \left(4\rho^2L_F^2+\frac{4}{n^2}\E\left[\norm{\nf{\w}{\z}}^2\right]\right)^{1/2}, \nonumber
	\end{align}
	where in the last inequality we use the following inequality by the symmetry between $\z$ and $\z'$
	\begin{align}\label{eq:s-stability-1}
		\E\left[\norm{\nf{\w}{\z}-\nf{\wi}{\z'}}^2\right]
		&\le 2\E\left[\norm{\nf{\w}{\z}}^2\right]
		+ 2\E\left[\norm{\nf{\wi}{\z'}}^2\right] \nonumber\\
		&= 4\E\left[\norm{\nf{\w}{\z}}^2\right]. 
	\end{align}
	Noting the distribution of $i_t$, we also have
	\begin{align}
		&\ \E\left[\norm{\nf{\w}{\z[i_t]}-\nf{\wi}{\z[i_t]'}}^2\right]
		= \E\left[\E_{i_t}\left[\norm{\nf{\w}{\z[i_t]}-\nf{\wi}{\z[i_t]'}}^2\right]\right] \nonumber\\
		=&\  
		\E\left[\frac{1}{n}\sum_{j\ne i}\norm{\nf{\w}{\z[j]}-\nf{\wi}{\z[j]}}^2
		+ \frac{1}{n}\norm{\nf{\w}{\z}-\nf{\wi}{\z'}}^2\right] \nonumber\\
		\le&\ 
		L_F^2\E\left[\norm{\w-\wi}^2\right] 
		+ \frac{4}{n}\E\left[\norm{\nf{\w}{\z}}^2\right], \nonumber
	\end{align}
	where we apply \cref{lem:s-weakly-properties,eq:s-stability-1} in the last inequality.
	Combining the above inequalities, we have
	\begin{align}
		\E\left[\norm{\w[t+1]-\wi[t+1]}^2\right]
		&\leq
		\left(1+\frac{2\eta C_2}{\sqrt{m}}+3\eta^2L_F^2\right)\E\left[\norm{\w-\wi}^2\right]
		+ \frac{12\eta^2}{n}\E\left[\norm{\nf{\w}{\z}}^2\right]
		+ 6\eta^2\rho^2L_F^2 \nonumber\\
		&\quad 
		+ 2\sqrt{2}\eta\left(\E\left[\norm{\w-\wi}^2\right]\right)^{1/2}\left(4\rho ^2L_F^2+\frac{4}{n^2}\E\left[\norm{\nf{\w}{\z}}^2\right]\right)^{1/2}. \nonumber
	\end{align}
	Summing up the above inequality over $t\in[k]$ and using $\w[1]=\wi[1]$, we have
	\begin{align}
		\E\left[\norm{\w[k+1]-\wi[k+1]}^2\right]
		&\leq
		\left(\frac{2\eta C_2}{\sqrt{m}}+3\eta^2L_F^2\right)\sum_{t=1}^{k}\E\left[\norm{\w-\wi}^2\right]
		+ \frac{12\eta^2}{n}\sum_{t=1}^{k}\E\left[\norm{\nf{\w}{\z}}^2\right]
		+ 6k\eta^2\rho^2L_F^2 \nonumber\\
		&\quad 
		+ 2\sqrt{2}\eta\sum_{t=1}^{k}\left(\E\left[\norm{\w-\wi}^2\right]\right)^{1/2}\left(4\rho ^2L_F^2+\frac{4}{n^2}\E\left[\norm{\nf{\w}{\z}}^2\right]\right)^{1/2} \nonumber\\
		&\leq
		\frac{1}{2}\max_{t\in[T]}\E\left[\norm{\w-\wi}^2\right]
		+ \frac{12\eta^2}{n}\sum_{t=1}^{k}\E\left[\norm{\nf{\w}{\z}}^2\right]
		+ 6k\eta^2\rho^2L_F^2 \nonumber\\
		&\quad 
		+ 2\sqrt{2}\eta\left(\max_{t\in[T]}\E\left[\norm{\w-\wi}^2\right]\right)^{1/2}\sum_{t=1}^{k}\left(4\rho ^2L_F^2+\frac{4}{n^2}\E\left[\norm{\nf{\w}{\z}}^2\right]\right)^{1/2}, \nonumber
	\end{align}
	where we use $\frac{2\eta C_2}{\sqrt{m}}+3\eta^2L_F^2\le \frac{1}{3T}$ since $m\ge36C_2^2T$ and $\eta\le\min\left\{\frac{1}{3\sqrt{2T}L_F},\frac{1}{4\sqrt{T}}\right\}$.
	Since the above inequality holds for any $k\in[T]$, we obtain
	\begin{align}
		\max_{t\in[T]}\E\left[\norm{\w-\wi}^2\right]
		&\leq \frac{18\eta^2}{n}\sum_{t=1}^{T}\E\left[\norm{\nf{\w}{\z}}^2\right]
		+ 9\eta^2\rho^2L_F^2T \nonumber\\
		&\quad
		+ 3\sqrt{2}\eta\left(\max_{t\in[T]}\E\left[\norm{\w-\wi}^2\right]\right)^{1/2}\sum_{t=1}^{T}\left(4\rho ^2L_F^2+\frac{4}{n^2}\E\left[\norm{\nf{\w}{\z}}^2\right]\right)^{1/2}, \nonumber
	\end{align}
	Solving the above quadratic inequality of $\left(\max_{t\in[T]}\E\left[\norm{\w-\wi}^2\right]\right)^{1/2}$ ($x^2\le a+bx, x\ge0$ implies $x^2\le 2a+b^2$), we obtain
	\begin{align}
		\max_{t\in[T]}\E\left[\norm{\w-\wi}^2\right]
		&\leq \frac{36\eta^2}{n}\sum_{t=1}^{T}\E\left[\norm{\nf{\w}{\z}}^2\right]
		+ 18\eta^2\rho^2L_F^2T \nonumber\\
		&\quad
		+ 18\eta^2\left[\sum_{t=1}^{T}\left(4\rho ^2L_F^2+\frac{4}{n^2}\E\left[\norm{\nf{\w}{\z}}^2\right]\right)^{1/2}\right]^2 \nonumber\\
		&\leq \frac{36\eta^2C_1^2}{n}\sum_{t=1}^{T}\E\left[F(\w;\z)\right]
		+ 18\eta^2\rho^2L_F^2T \nonumber\\
		&\quad
		+ 18\eta^2\left[\sum_{t=1}^{T}\left(4\rho ^2L_F^2+\frac{4C_1^2}{n^2}\E\left[F(\w;\z)\right]\right)^{1/2}\right]^2 \nonumber\\
		&\leq \frac{36\eta^2C_1^2}{n}\left(1+\frac{2T}{n}\right)\sum_{t=1}^{T}\E\left[F(\w;\z)\right]
		+ 90\eta^2\rho ^2L_F^2T^2, \nonumber
	\end{align}
	where we apply \cref{lem:s-weakly-properties} in the second inequality and Cauchy-Schwarz inequality in the last one.
	Summing up the above inequality over $i\in[n]$, we obtain the results.
\end{proof}

\begin{proof}[Proof of \cref{thm:ss-gen}]
	Combining \cref{thm:s-conv,thm:ss-stability,thm:ss-on-ave}, we obtain
	\begin{align}
		\E\left[\F[\D](\w)\right] 
		&\le  \frac{18\eta^2C_1^2(1+L_F)}{n}\left(1+\frac{2T}{n}\right)\frac{4\norm{\w[1]-\w[]}^2}{\eta} 
		+ 45(1+L_F)\eta^2\rho^2L_F^2T^2
		+ \left(1+\frac{C_1^2}{2}\right)\E\left[\F(\w)\right]. \nonumber
	\end{align}
	Summing up the above inequality over $t\in\{0\}\cup[T-1]$ and combining it with \cref{thm:s-conv}, we obtain
	\begin{align}
		\frac{1}{T}\sum_{t=1}^{T}\E\left[\F[\D](\w)\right] 
		&\le 
		\frac{18\eta^2C_1^2\left(1+L_F\right)}{n}\left(1+\frac{2T}{n}\right)\frac{4\norm{\w[1]-\w[]}^2}{\eta} \nonumber\\
		&\quad
		+ 45\left(1+L_F\right)\eta^2\rho^2L_F^2T^2 
		+ \frac{1}{T}\left(1+\frac{C_1^2}{2}\right)\sum_{t=1}^{T}\E\left[\F(\w)\right] \nonumber\\
		&\le 
		\left( \frac{18\eta^2C_1^2\left(1+L_F\right)}{n}\left(1+\frac{2T}{n}\right)
		+ \frac{1}{T}\left(1+\frac{C_1^2}{2}\right)\right)\frac{4\norm{\w[1]-\w[]}^2}{\eta} \nonumber\\
		&\quad
		+ 45\left(1+L_F\right) \eta^2\rho^2L_F^2T^2 \nonumber\\
		&\le
		\cO\left(
		\left(1+\frac{1}{n}+\frac{T}{n^2}\right)\frac{\norm{\w[1]-\w[]}^2}{\eta T} 
		+ \frac{\norm{\w[1]-\w[]}^2}{T}\right), \nonumber
	\end{align} 
	where we use $\eta^2L_F^2\le\frac{1}{9T}$ and $\rho\le\frac{2\norm{\w[1]-\w[]}}{T}$ in the last inequality. 
	
	Finally, if \cref{as:nn-realizability} holds with sufficiently small $\epsilon\le1/T$, we may take $\w[]$ as $\w[\epsilon]$, for which $\F(\w[\epsilon])\le\epsilon$ and $\norm{\w[1]-\w[\epsilon]}\le g(\epsilon)$. The step size $\eta \le \frac{\norm{\w[1]-\w[\epsilon]}^2}{8}$ then implies 
	\begin{align}
		8\eta 	\sum_{t=1}^{T}F(\w[\epsilon];\z[i_t]) \le T \epsilon \norm{\w[1]-\w[\epsilon]}^2  \le\norm{\w[1]-\w[\epsilon]}^2, \nonumber
	\end{align}
	so \cref{eq:s-w} is satisfied. Substituting these estimates into the previous bound yields
	\begin{align}
		\E\left[\F[\D]\left(\bar{\bm{w}}_T\right)\right]
		&\le
		\cO\left(\left(1+\frac{1}{n}+\frac{T}{n^2}\right)\frac{g(\epsilon)^2}{T\eta} + \frac{g(\epsilon)^2}{T}\right). \qedhere \nonumber
	\end{align}
\end{proof}


\section*{Funding}
National Natural Science Foundation of China (12471096); National Key Research and Development Program of China (2021YFA1003500).

\bibliographystyle{plain}
\bibliography{ref}


\end{document}